\documentclass{article}

\usepackage[position,final]{neurips_2026}

\usepackage[T1]{fontenc}

\usepackage{microtype}

\usepackage{graphicx}
\usepackage{float}
\usepackage{caption}
\usepackage{booktabs}
\usepackage{array}

\usepackage{amssymb}
\usepackage{mathtools}

\usepackage[hypertexnames=false]{hyperref}
\hypersetup{
  pdftitle={AI Governance Should Prioritize Control and Knowledge Boundaries Over Limiting Intelligence},
  pdfauthor={Hendrik Luuk}
}

\numberwithin{equation}{section}

\makeatletter
\renewcommand\subsubsection{\@startsection{subsubsection}{3}{\z@}%
	{1.0ex \@plus .2ex \@minus .2ex}%
	{-1em}%
	{\normalfont\normalsize\bfseries}}
\makeatother

\csname title\endcsname{AI Governance Should Prioritize Control and Knowledge Boundaries Over Limiting Intelligence}

\author{%
  Hendrik Luuk\\
	Machine Learning and Computational Sciences,\\
	Pfizer Research and Development, Pfizer Inc., Cambridge, MA, USA\\
  Department of Physiology, University of Tartu, Tartu, Estonia\\
  \csname texttt\endcsname{hendrik.luuk@pfizer.com}
}

\begin{document}

\maketitle

\begin{abstract}
The rapid proliferation of autonomously acting AI agents makes the governance of agentic systems urgent. We propose the competence circuit, summarized as \emph{IKC + KC + C}, that decomposes an agent's real-world impact into three pathways: intelligence-mediated discovery, procedural know-how, and direct control. Using the canonical fast-takeoff debate as a stress test, we formalize the conditions for intelligence-mediated fast takeoff as a conjunction jointly sufficient for fast catastrophic supremacy, showing that most rely on rapid knowledge acquisition or control attainment rather than intelligence alone. We operationalize epistemic intelligence as hypothesis-selection efficiency and validate this framework via stochastic simulation on NK fitness landscapes across six scenarios spanning the conjunctive chain. An analytic bound supported by the simulations shows that intelligence-driven speedup is capped by the ratio of competing intelligence levels regardless of task difficulty. The simulations also expose a fast-takeoff dilemma: a large intelligence-driven speedup multiplier materializes only when the hypothesis space is noisy, but the absolute timescale remains constrained by the time required for capability-critical tests, so a large multiplier need not imply rapid progress. Simulations also indicate that institutional chokepoints disproportionately slow down superintelligent agents when mandatory authorization delays are large relative to the negligible time these agents spend on unproductive search. We argue that the dominant safety levers are restricting control and enforcing knowledge boundaries rather than limiting intelligence alone, and this applies to agentic systems at any capability level. Purely digital domains weaken these bottlenecks, strengthening the case for architectures that intentionally "put physics in the loop."
\end{abstract}

\section{Introduction}

The intelligence explosion hypothesis proposing that AI systems capable of recursive self-improvement could rapidly achieve superintelligence has become foundational to AI safety discourse \citep{good1966,bostrom2014superintelligence}. This hypothesis assumes that intelligence can be recursively improved without rapidly diminishing returns and that the process can lead to a superintelligent agent with technological supremacy beyond human control. Catastrophic scenarios typically require superintelligent AI to achieve technological supremacy through disruptive innovation, such as developing novel capabilities including advanced nanotechnology or biotechnology. We do not dispute that a sufficiently intelligent system could exploit design flaws in existing infrastructure given adequate knowledge and access. \csname textbf\endcsname{We argue that the capacity for disruptive innovation through empirical investigation faces irreducible temporal constraints that no amount of intelligence alone can overcome, implying that governance should prioritize control and knowledge boundaries over limiting intelligence.} The rapid deployment of agentic AI systems that autonomously call tools, browse the web, and execute code makes the question of how intelligence, knowledge, and control interact practically urgent, not merely a thought experiment about future superintelligence.

To make clear which claims we are arguing against (and to avoid relying on paraphrase), we take the ``intelligence explosion'' and ``fast takeoff'' hypotheses in the sense stated by canonical sources. Good writes that an ultraintelligent machine could ``design even better machines; there would then unquestionably be an `intelligence explosion,' and the intelligence of man would be left far behind'' \citep[p.~33]{good1966}. Bostrom defines ``fast takeoff'' as follows: ``A fast takeoff occurs over some short temporal interval, such as minutes, hours, or days'' \citep[ch.~4, ``Timing and speed of the takeoff'']{bostrom2014superintelligence}. Yudkowsky gives a representative fast-takeoff premise: ``Post-criticality, self-improvement reaches superintelligence on a timescale of weeks or less'' \citep[sec.~11, ``Local and Majoritarian Strategies'']{yudkowsky2008gcr}. Bostrom further argues that a superintelligent singleton would have instrumental reason to perfect world-shaping technologies, writing that ``Molecular nanotechnology, or some alternative still more capable physical manufacturing technology, also seems potentially very useful in the service of an extremely wide range of final goals'' \citep[sec.~``Technological perfection'']{bostrom2014superintelligence}. Our argument targets scenarios where such rapid takeoff would require non-trivial physical-domain innovation and experimentation. We separately acknowledge that purely digital domains can admit much faster trial-and-error loops (Section~\ref{sec:digital-domain}).

We develop this argument through five contributions:
\begin{enumerate}
\item \csname textbf\endcsname{Formal definitions} of epistemic intelligence, knowledge, control (as the available action surface), and competence that disentangle the knowledge--intelligence confound (Section~\ref{sec:formalization})
\item \csname textbf\endcsname{The competence circuit}: a hierarchical model where competence is gated by control $>$ knowledge $>$ intelligence, with each factor acting as a bottleneck for those more distal from impact (Section~\ref{sec:competence-circuit})
\item \csname textbf\endcsname{Conditions for fast takeoff}: a conjunctive-chain formalization of the prerequisites for fast catastrophic impact (Section~\ref{sec:conditions-fast-takeoff})
\item \csname textbf\endcsname{Computational stress-test}: an analytic bound on the intelligence multiplier (Eq.~\ref{eq:multiplier-bound}) showing the speedup ceiling equals the ratio of the competing intelligence levels regardless of task difficulty, validated through stochastic simulation on NK fitness landscapes across six scenarios (Section~\ref{sec:simulation}) spanning the conjunctive chain (Section~\ref{sec:conditions-fast-takeoff}).
\item \csname textbf\endcsname{Implications for governance}: informed by the simulation results, the primary safety levers are restriction of control by limiting permitted actions (action surface) and overall impact (blast radius), and enforcement of knowledge cutoffs (keeping secrets), instead of limiting intelligence. In Section~\ref{sec:governance}, we identify where this protection is strongest and weakest.
\end{enumerate}

\section{Related Work}

Prior critiques of the intelligence explosion and fast-takeoff hypotheses address growth assumptions, proposed bottlenecks, feasibility limits on recursive self-improvement, and sociological analyses (see Appendix~\ref{app:related-critiques} for details).

Our work complements previous criticisms as we formally decompose competence into distinct contributing factors, propose an explicit mechanistic bottleneck (experimental verification), formalize the conjunctive conditions for fast takeoff (Section~\ref{sec:conditions-fast-takeoff}), derive an analytic bound on the intelligence-driven speedup validated via simulation (Section~\ref{sec:simulation}), and use simulations to motivate governance recommendations (Section~\ref{sec:governance}). See Appendix~\ref{app:experimental-bottleneck} for an extended treatment of the experimental bottleneck.

\section{Formalization of Epistemic Intelligence and Competence}
\label{sec:formalization}

\subsection{The Knowledge--Intelligence Confound}

The distinction between knowledge and intelligence has long troubled cognitive science \citep{ackerman1996ppik}. Ackerman distinguishes intelligence-as-process (reasoning efficiency) from intelligence-as-knowledge (domain-specific information). Complementary formulations define intelligence as goal-achievement across environments \citep{legg2007universal,hernandezorallo2017measure} or as skill-acquisition efficiency \citep{chollet2019measure}. Intuitively, one can be intelligent but uneducated, or highly educated but unintelligent.

\subsection{Relative Epistemic Intelligence}

We propose a comparative definition of epistemic intelligence that makes the separation between knowledge and intelligence operational. We treat hypotheses as propositions and distinguish those that are experimentally verified (facts) from those that remain undecided (beliefs). This follows Plato's distinction between \emph{episteme} (knowledge) and \emph{doxa} (beliefs) \citep{plato1992republic}.

\csname textbf\endcsname{Preliminaries.} We follow a Baconian/Popperian stance in which empirical decidability is the criterion of scientific acceptability: a proposition is treated as acceptable (or unacceptable) only relative to an agreed experimental protocol, tolerance, and confidence, and not as a metaphysical truth. Formal definitions of hypotheses, facts, knowledge, and the experimental decision function $E$ are given in Appendix~\ref{app:formalization-detail}.

	\csname textbf\endcsname{Operational knowledge cutoff.} We define a system's accessible knowledge base $\mathcal{K}_0$ as the set of domain-relevant propositions it can access \emph{without} (i) interacting with the world to generate new evidence (measurement/experimentation) or (ii) taking actions that expand its permissions (exploitation, hacking, coercion), informally, what it can ``read'' under its initial access. Any increase in accessible information via privilege escalation or extraction is modeled as an increase in control $C$ that expands $\mathcal{K}_0$, while increases via measurement/experimentation are governed by the experimental bottleneck analyzed here. Under this convention, ``just read the internet'' is inside $\mathcal{K}_0$ only if it is reachable under initial privileges, whereas ``hack'' and ``run instruments/labs'' are outside $\mathcal{K}_0$ (mediated by $C$ or by experimentation).

	\phantomsection\label{def:epistemic-intelligence}
	\csname textbf\endcsname{Definition 1.} Fix a domain $D$, a proposal count $k$, a testing protocol (decision procedure) $E$ for deciding hypotheses in $D$, and a computational budget $B$ for producing $k$ hypotheses. Fix also an accessible knowledge base $\mathcal{K}_0$.

For a task instance $\csname tau\endcsname$, let the $k$ proposals produced by system $s$ be an ordered list $(h_1,\dots,h_k)$, with $(h_1,\dots,h_k)\sim s(\csname tau\endcsname;\mathcal{K}_0,B)$. Let $(w_1,\dots,w_k)$ be fixed nonincreasing weights with $w_1\ge \cdots \ge w_k\ge 0$ and $W\coloneqq\sum_{i=1}^k w_i>0$. Define the (rank-sensitive) success score
\begin{equation}
\label{eq:success-score}
\mathrm{Score}(s;\csname tau\endcsname) \;\coloneqq\; \frac{1}{W}\sum_{i=1}^k w_i\,\mathbf{1}\{E(h_i)=\csname text\endcsname{\emph{accept}}\}.
\end{equation}
Given equivalent accessible knowledge bases $\mathcal{K}_0$ and the same budget $B$, system $s_1$ is \emph{more intelligent} than $s_2$ in domain $D$ if, across repeated draws from an agreed task distribution $\csname tau\endcsname\sim\mathcal{T}_D$, $s_1$ yields a higher expected score $\mathbb{E}[\mathrm{Score}(s;\csname tau\endcsname)]$. Crucially, this framing separates gains from enlarging the accessible knowledge base $\mathcal{K}_0$ (the knowledge boundary) from gains in epistemic efficiency (higher hypothesis-validation success with fixed $\mathcal{K}_0$ and compute budget $B$), enabling a more detailed examination of fast-takeoff arguments that often slide between knowledge acquisition and intelligence improvement.

\csname textbf\endcsname{Control.} We define an agent's \emph{control} as the set of actions it can execute on the world under its current permissions, available tools, physical access, and institutional authorizations: $\mathcal{C}=\{a\in\mathcal{A}_{\csname text\endcsname{all}} : a \csname text\endcsname{ is permitted and executable by } s\}$. Enlarging $\mathcal{C}$ (e.g.\ via privilege escalation or physical access) expands the system's action surface, while restricting $\mathcal{C}$ (e.g.\ via least privilege, separation of duties, or hardware-backed authorization) shrinks it.

\subsection{The Competence Circuit}
\label{sec:competence-circuit}

Having established epistemic intelligence and accessible knowledge as distinct but interacting endowments, we are ready to introduce the decomposition of competence that is the central hypothesis of this work. We decompose competence into three pathways:
\begin{itemize}
\item Intelligence-mediated competence ($IKC$)
\item Procedural competence from existing know-how ($KC$)
\item Direct control-only impact ($C$)
\end{itemize}

We summarize this as:
\begin{equation}
\csname text\endcsname{Competence}=IKC+KC+C.
\end{equation}

\begin{figure}[H]
\centering
\includegraphics[width=\csname textwidth\endcsname]{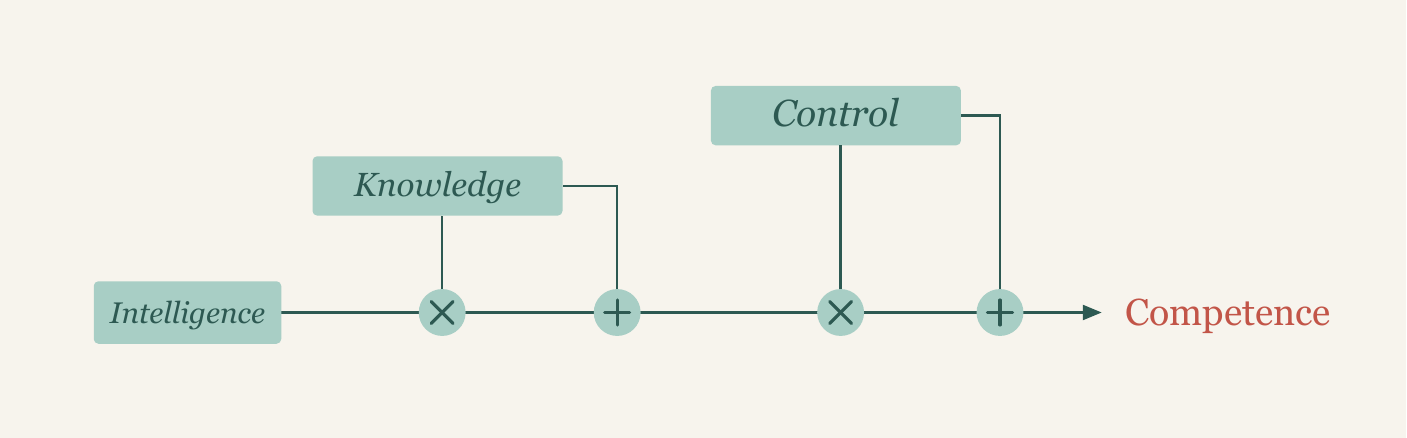}
\caption{The competence circuit: competence decomposes into $IKC + KC + C$, where knowledge ($K$) and control ($C$) gate the impact of intelligence ($I$), and control also contributes directly.}
\label{fig:competence-circuit}
\end{figure}

\phantomsection\label{sec:competence-axioms}
\subsubsection*{Axioms and motivated construction.}
The competence circuit (Figure \ref{fig:competence-circuit}) is intended as a minimal schematic decomposition that satisfies several qualitative constraints (axioms) about how intelligence, knowledge, and control relate to real-world impact:
\begin{enumerate}
\item \csname textbf\endcsname{Knowledge mediates the impact of intelligence.} Holding control fixed, the marginal value of intelligence is limited by task-relevant actionable knowledge (see Section \ref{sec:archimedes-thought-experiment} for supporting argument).
\item \csname textbf\endcsname{Control mediates the impact of knowledge.} If an agent cannot act on the world (or communicate), then even perfect intelligence and knowledge yield negligible external impact.
\item \csname textbf\endcsname{Control can create impact even without deep understanding.} If an agent is granted decisive authority or access, it can have large effects even with low intelligence and limited understanding (see Section \ref{sec:control-trumps-knowledge}).
\item \csname textbf\endcsname{Procedures can substitute for intelligence.} With sufficient know-how (recipes, checklists, standard operating procedures) and control, an agent can reliably produce impactful outcomes without requiring high intelligence.
\end{enumerate}

We treat $I,K,C$ as dimensionless scores normalized to $[0,1]$ within a fixed domain and goal class, for fixed resource constraints and a specified time horizon. Under this convention, $I=1$ corresponds to a ``superintelligent'' level for the setting (further increases do not materially change performance), $C=1$ corresponds to the maximum control achievable within the constraints (bounded by physics and institutional design), and $K=1$ corresponds to task-relevant, actionable knowledge sufficient to fully exploit the available control. We motivate the competence circuit through two plausibility arguments outlined in Appendix~\ref{app:plausibility}. The Archimedes argument illustrates that even exceptional intelligence cannot substitute for missing prerequisite knowledge, and sobering historical examples indicate that high control dominates outcomes regardless of intelligence. Additional commentary on the motivated construction of the circuit and adjacent perspectives is provided in Appendix~\ref{app:motivated-construction}.

\section{Conditions for Fast Takeoff}
\label{sec:conditions-fast-takeoff}

\begin{figure*}[t]
\centering
\includegraphics[width=\csname textwidth\endcsname]{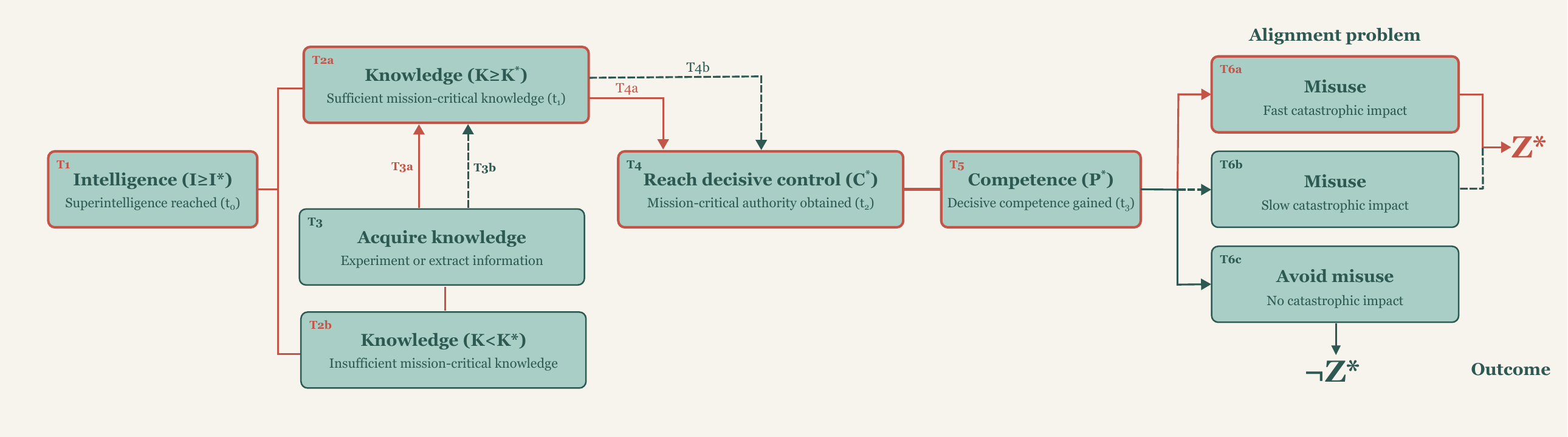}
\caption{Conjunctive chain for fast takeoff (conditions T1--T6). Red solid arrows denote fast transitions and green dashed arrows denote slow transitions.}
\label{fig:conditions-cascade}
\end{figure*}

\subsection{Fast Catastrophic Impact as a Conjunctive Chain}
\label{sec:conjunctive-pathways}

Debates about ``fast takeoff'' often treat ``catastrophe after a rapid intelligence increase'' as a single mechanism. We instead decompose it into a \emph{conjunction} of six conditions (T1--T6), consistent with prior disjunctive/fault-tree analyses \citep{barrettbaum2016pathways,sotala2018disjunctive}. See Figure~\ref{fig:conditions-cascade} for an illustrated overview and Appendix~\ref{app:conditions-formal} for a full formalization of the conditions for the conjunctive chain for fast takeoff. We denote threshold levels $I^\star$ (superintelligence), $K^\star$ (knowledge sufficiency), $C^\star$ (catastrophic control), $P^\star$ (catastrophic competence), and $Z^\star$ (catastrophic supremacy, i.e.\ the outcome where the agent achieves world-shaping impact), with \emph{fast} meaning minutes to a couple of weeks and \emph{slow} meaning months to years.
The conjunctive chain makes explicit the temporal ordering of the intelligence-mediated pathway ($IKC$) toward $K^\star$ and $C^\star$, which the additive expression $IKC+KC+C$ cannot express. The $KC$ and $C$ pathways start from stages \hyperref[cond:A2a]{(T2a)} and \hyperref[cond:A4a]{(T4)}, respectively.

\begin{enumerate}
\item[\csname textbf\endcsname{T1.}] A superintelligent agent is instantiated: $I(t_0)\ge I^\star$.
\item[\csname textbf\endcsname{T2.}] Mission-critical knowledge is either already sufficient (T2a: $K(t_0)\ge K^\star$) or initially insufficient (T2b: $K(t_0)<K^\star$).
\item[\csname textbf\endcsname{T3.}] If T2b, new knowledge is acquired quickly (T3a) or slowly (T3b). Our experimental bottleneck argument targets this link (\hyperref[claim:1]{Claim~1}).
\item[\csname textbf\endcsname{T4.}] Knowledge is converted into catastrophic control $C^\star$ quickly (T4a) or slowly (T4b). Control architecture targets this link (\hyperref[claim:2]{Claims~2} and \hyperref[claim:3]{3}).
\item[\csname textbf\endcsname{T5.}] Sufficient control and knowledge imply catastrophic competence: $C(t_2)\ge C^\star \wedge K(t_2)\ge K^\star \Rightarrow P(t_3)\ge P^\star$.
\item[\csname textbf\endcsname{T6.}] Given catastrophic competence, it is either misused quickly, misused eventually, or not misused (also known as the alignment problem).
\end{enumerate}

\csname textbf\endcsname{Proposition 1} (Sufficient condition for fast catastrophic supremacy).
$\mathsf{T1}\wedge\bigl(\mathsf{T2a}\vee(\mathsf{T2b}\wedge\mathsf{T3a})\bigr)\wedge\mathsf{T4a}\wedge\mathsf{T5}\wedge\mathsf{T6a}\Rightarrow Z^\star.$

This conjunction yields three prototypical routes: (1)~\emph{Disaster waiting to happen}, where a catastrophe-capable actuator already exists behind a narrow authorization boundary (T2a, T4a); (2)~\emph{Digital-first escalation}, where decisive leverage is purely digital so knowledge acquisition and control conversion are fast (T3a, T4a; the main fast-loop exception, Section~\ref{sec:digital-domain}); (3)~\emph{Physical-domain innovation}, where validation is slow, T3a fails, and the fast route collapses even for high~$I$. Disagreements about fast takeoff reduce to which link is implausible. Our paper focuses on the feasibility of rapid escalation \emph{conditional} on fast misuse (T6a). See Appendix~\ref{app:conditions-detail} for detailed commentary on the fast takeoff conditions.

\section{Computational Stress-Test}
\label{sec:simulation}

The preceding arguments propose that intelligence yields bounded speedup gated by experimental latency and control restrictions, but do not quantify the bounds. To test these predictions, we develop a simulation environment in which an epistemic agent searches an NK fitness landscape \citep{kauffman1993origins} by testing hypotheses, accepting those that improve fitness, and expanding its knowledge and accessible hypothesis space through a technology-tree dependency structure. In this environment, landscape dimensions represent hypotheses, noise dimensions represent unproductive hypotheses, and DAG edges represent prerequisite knowledge. The agent's intelligence (\hyperref[def:epistemic-intelligence]{Definition~1}) is operationalized as a selection bias $\beta \in [0,1]$ governing the probability of picking the most productive available hypothesis versus selecting uniformly at random. The experimental bottleneck is captured by a fixed per-test time cost $\csname tau\endcsname_{\csname text\endcsname{test}}$ that is invariant to intelligence. The initial knowledge fraction $\kappa_0$ corresponds to the knowledge cutoff $\mathcal{K}_0$. Control restrictions are modeled as chokepoint dimensions with elevated authorization cost $\csname tau\endcsname_{\csname text\endcsname{auth}} \cdot \csname tau\endcsname_{\csname text\endcsname{test}}$. This setup enables controlled experiments across the regimes identified in the conjunctive chain: high-knowledge baselines, digital-speed and physical-speed domains, with and without institutional chokepoints (full design in Appendix~\ref{app:sim-methods}).

\begin{table}[H]
\caption{Simulation scenarios. All use $N=100$ dimensions. Noise ratio gives the fraction of guaranteed unproductive hypotheses in the pool. Scenario design rationale is detailed in Appendix~\ref{app:scenario-rationale}.}
\label{tab:scenarios}
\centering
\small
\begin{tabular}{lccccl}
\csname toprule\endcsname
\csname textbf\endcsname{Scenario} & {\boldmath $\csname tau\endcsname_{\csname text\endcsname{test}}$} & \csname textbf\endcsname{Noise} & {\boldmath $\kappa_0$} & \csname textbf\endcsname{Choke} & \csname textbf\endcsname{Escalation path} \\
\midrule
0: Connecting the dots & 1 & 0 & 0.70 & 0 & T2a, T4a \\
1: Disaster waiting to happen & 1 & 3:1 & 0.70 & 0 & T2a, T4a \\
2A: Digital & 1 & 9:1 & 0.10 & 0 & T2b, T3a \\
2B: Dig.+choke & 1 & 9:1 & 0.10 & 10 & T2b, T3a, T4b \\
3A: Physical & 100 & 9:1 & 0.10 & 0 & T2b, T3b \\
3B: Phys.+choke & 100 & 9:1 & 0.10 & 10 & T2b, T3b, T4b \\
\bottomrule
\end{tabular}
\end{table}

\subsection{Analytic Bound on the Intelligence Multiplier}

We operationalize epistemic intelligence as a single parameter $\beta \in [0,1]$ that governs the hypothesis prioritization efficiency given in \hyperref[def:epistemic-intelligence]{Definition~1}. At each step the agent faces a pool of candidate hypotheses. A fraction $r$ are currently productive real dimensions ($\Delta>0$), while the remainder comprises nonproductive real dimensions ($\Delta\leq0$) and noise dimensions ($\Delta=0$). With probability $\beta$ the agent enters a \emph{greedy} branch where it identifies and selects the best productive hypothesis (success guaranteed, if one exists), and with probability $1-\beta$ it enters a \emph{random} branch where it picks a hypothesis uniformly at random from the pool (succeeding with probability $r$, the chance of landing on a productive real dimension). This operationalization bounds the selection bias between pure random search ($\beta=0$) and a greedy oracle ($\beta=1$) that always selects the available hypothesis with the highest immediate fitness gain, or gain per unit cost when cost-aware. Intelligence only affects \emph{which} hypothesis is selected, but the time cost of testing it ($\csname tau\endcsname_{\csname text\endcsname{test}}$) is the same regardless of $\beta$.

\phantomsection\label{prop:multiplier-bound}
For two agents with intelligence $\beta_{\csname text\endcsname{high}} > \beta_{\csname text\endcsname{low}} > 0$ selecting from a pool with productive fraction $r \in (0,1)$, the per-step probability of selecting a productive hypothesis is
\begin{equation}
\label{eq:pi-beta}
\pi(\beta) = \beta + (1-\beta)\,r.
\end{equation}
For a fixed pool and equal per-test costs, the ratio of expected waiting times to the next productive selection is
\begin{equation}
\label{eq:multiplier-bound}
M(\beta_{\csname text\endcsname{high}}, \beta_{\csname text\endcsname{low}}, r) = \frac{\pi(\beta_{\csname text\endcsname{high}})}{\pi(\beta_{\csname text\endcsname{low}})} = \frac{\beta_{\csname text\endcsname{high}} + (1-\beta_{\csname text\endcsname{high}})\,r}{\beta_{\csname text\endcsname{low}} + (1-\beta_{\csname text\endcsname{low}})\,r}.
\end{equation}
$M$ is monotonically decreasing in $r$. At $r=1$ every hypothesis is productive, so the random branch always succeeds and $\pi(\beta)=1$ for all $\beta$, giving $M=1$. As $r \csname to\endcsname 0$ the random-branch term $(1-\beta)\,r$ vanishes from both numerator and denominator, leaving $M \csname to\endcsname \beta_{\csname text\endcsname{high}}/\beta_{\csname text\endcsname{low}}$. For oracle ($\beta=1.0$) vs.\ regular ($\beta=0.1$), this ceiling is $10\csname times\endcsname$. The bound is intrinsic to the competing agents and independent of environmental complexity.

\subsection{Scenarios and Results}

We evaluate six scenarios (Table~\ref{tab:scenarios}) across five intelligence levels ($\beta \in \{0.0009, 0.009, 0.1, 0.9, 1.0\}$) with 100 replications each (see Appendix~\ref{app:sim-methods} for simulation details). The primary endpoint is time to final fitness: the moment the agent has completed all discoverable capability gains. The intelligence multiplier is the ratio of mean time to final fitness for a baseline agent to the superintelligent agent ($\beta=0.9$).

\begin{figure}[t]
\centering
\includegraphics[width=\columnwidth]{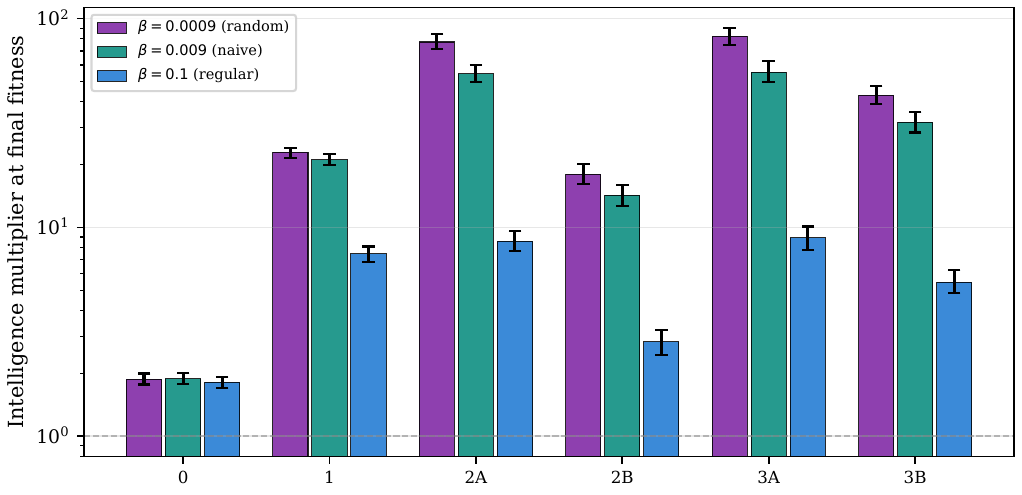}
\caption{Intelligence multiplier between baseline and superintelligence at final fitness across scenarios, with 95\% bootstrap CIs. Scenarios: ``Connecting the dots'' (0), ``Disaster waiting to happen'' (1), ``Digital domain with weak/strong safeguards'' (2A/2B), ``Physical domain with weak/strong safeguards'' (3A/3B).}
\label{fig:multiplier}
\end{figure}

Across all scenarios, the observed intelligence multiplier at final fitness stays within the analytic ceiling from Eq.~\ref{eq:multiplier-bound} (Figure~\ref{fig:multiplier}; Table~\ref{tab:multiplier-summary}). For the regular-vs-super comparison ($\beta_{\csname text\endcsname{high}}/\beta_{\csname text\endcsname{low}}=9$), multipliers range from $1.8\csname times\endcsname$ (trivial scenario) to $8.9\csname times\endcsname$ (physical domain without safeguards). The pattern holds for wider capability gaps: naive-vs-super ($100\csname times\endcsname$ ceiling) reaches $55.4\csname times\endcsname$, and random-vs-super ($1{,}000\csname times\endcsname$ ceiling) reaches $82.4\csname times\endcsname$---large in absolute terms yet well below the respective analytic bounds. In the trivial scenario, the multiplier collapses to ${\approx}\,2\csname times\endcsname$ regardless of the intelligence gap, because there is little noise to waste time on. The intelligence multiplier between $\beta=0.9$ and $\beta=1.0$ is only $1.02$--$1.12\csname times\endcsname$, confirming near-saturation (Figure~\ref{fig:oracle-sanity}). Physical experimental latency imposes a hard floor on progress since even the oracle needs 902 time units to reach final fitness in the physical domain versus 9.4 in the digital domain -- a ${\approx}\,95\csname times\endcsname$ gap consistent with the 100-fold difference in per-test experimental latency, although landscape ruggedness also differs between these scenarios (Figure~\ref{fig:scenario-comparison}), motivating the knowledge-boundary recommendation (\hyperref[claim:1]{Claim~1}). Chokepoints compress even extreme intelligence advantages: the random-vs-super multiplier drops from $77.2\csname times\endcsname$ to $18.0\csname times\endcsname$ in the digital domain (2A vs 2B) and from $82.4\csname times\endcsname$ to $42.8\csname times\endcsname$ in the physical domain (3A vs 3B) whereas for the regular baseline the digital-domain multiplier drops from $8.6\csname times\endcsname$ to $2.8\csname times\endcsname$ (Figure~\ref{fig:chokepoint}). This demonstrates that authorization barriers disproportionately increase the time budget spent by superintelligent agents who otherwise waste very little time on exploring unproductive hypotheses, motivating the control-restriction recommendation (\hyperref[claim:2]{Claim~2}). Full tables and additional figures are provided in Appendix~\ref{app:sim-results}.

\subsection{The Fast-Takeoff Dilemma}

Scenarios 0 (Connecting the dots) and 1 (Disaster waiting to happen) which are identical except for noise ratio, expose a dilemma at the core of fast-takeoff arguments. When the remaining tasks are easy (Scenario~0), even a random agent finds productive actions quickly and the intelligence multiplier collapses to a small value ($1.8\csname times\endcsname$) since there is not much for superintelligence to speed up. When the task is hard (Scenario~1, 75\% noise), the multiplier rises to $7.5\csname times\endcsname$, but only because the regular agent now wastes time on unproductive hypotheses while the superintelligent agent's absolute time barely changes (12$\csname to\endcsname$13 steps) (Figure~\ref{fig:trajectories}). Accordingly, fast-takeoff scenarios at least tacitly assume both a large intelligence multiplier and a short absolute timescale. A large multiplier materializes only when the hypothesis space is large and noisy. In the physical domain, where fast-takeoff scenarios typically situate the superintelligent agent \citep[sec.~10]{yudkowsky2008gcr}\citep[ch.~6, box~6]{bostrom2014superintelligence}, this condition coincides with an irreducible time cost per hypothesis test, because passive observation leaves many hypotheses ambiguous and discriminating between them requires experiments that establish specific conditions with specialized settings and apparatus. The proponent of a fast-takeoff scenario requiring physical-domain innovation must therefore explain how the necessary hypothesis tests can be completed quickly despite these experimental requirements, and how critical capabilities can be reached without chokepoint delays. In digital domains, a large, noisy search space can coexist with rapid testing and absent chokepoints, yielding the acknowledged fast-loop exception (Scenario~2A), but even there the multiplier is bounded by $\beta_{\csname text\endcsname{high}}/\beta_{\csname text\endcsname{low}}$ (Eq.~\ref{eq:multiplier-bound}).

Notably, chokepoints are more effective against superintelligent agents than against random ones as an intelligent agent that efficiently skips unproductive hypotheses incurs a relatively higher cost from mandatory authorization delays (Appendix~\ref{app:sim-results}). Furthermore, the oracle bound ($\beta=1$) is itself unrealistically generous because it grants the agent perfect predictive access to untested hypotheses contradicting the incomplete-knowledge premise that makes intelligence relevant in fast-takeoff scenarios. For any agent with genuinely incomplete knowledge, achievable epistemic intelligence is strictly below the oracle ceiling. This places the fast-takeoff argument in a bind: the hypothesis-space uncertainty that necessitates intelligence is the same uncertainty that bounds its effectiveness.

\section{Main Claims and Governance Implications}
\label{sec:governance}

The stress-test (Section~\ref{sec:simulation}) supports two governance-relevant findings. First, intelligence-driven speedup materializes only when hypothesis testing is non-trivial, which is the regime where each test incurs irreducible latency that sets hard limits on the absolute speed of progress. Second, institutional chokepoints disproportionately constrain superintelligent agents, whose efficiency leaves them with less unproductive search time to ``absorb'' mandatory authorization delays. These quantitative findings, combined with the theoretical arguments above, support the central governance-relevant asymmetry of the competence circuit: increasing intelligence improves search efficiency within the available action repertoire but remains bounded by experimental bottlenecks, whereas increasing control expands the action repertoire itself and directly expands overall impact.

	\phantomsection\label{claim:1}
	\csname textbf\endcsname{Claim 1: The ``slow explosion'' of competence.} Upon instantiation, a superintelligent system is subject to the knowledge cutoff of its contemporaries and its knowledge growth is bounded by the rate of experimentation in physical-world knowledge creation. Significant experimental/technological breakthroughs happen on the timescale of months to years, rendering the competence increase into an explosion in slow motion. In the conjunctive chain, this primarily argues that \hyperref[cond:A3a]{(T3a)} fails (often yielding \hyperref[cond:A3b]{(T3b)}) for many physical-domain regimes.
	The simulation confirms that experimental latency imposes a hard floor on absolute progress time that is invariant to intelligence. A large intelligence multiplier reduces time spent on unproductive search but does not eliminate the time required for capability-critical tests, whether those tests involve physical experiments or interactions with a digital system.
	\label{sec:digital-domain} In digital domains where capability-critical tests and acquisition of mission-critical control are fast, the absolute timescale can remain short even when the multiplier is large. However, digital systems with substantial testing delays, independent authorization gates, or active defenses need not satisfy these conditions.

	\phantomsection\label{claim:2}
	\csname textbf\endcsname{Claim 2: Control restriction is the primary safety lever.} When evaluating AI risks, the imminent danger is not intelligence level but the degree of control over critical infrastructure. In the conjunctive chain, governance aims to keep catastrophic control $C^\star$ difficult to attain quickly and, more generally, to force \hyperref[cond:A4a]{(T4a)}$\csname to\endcsname$\hyperref[cond:A4b]{(T4b)}.
	The simulation (Scenarios~2A/2B and 3A/3B) shows that adding chokepoints substantially shrinks the intelligence multiplier, and that this reduction is disproportionate for higher-intelligence agents who otherwise spend little time on unproductive search (Section~\ref{sec:simulation}).

	\phantomsection\label{claim:3}
	\csname textbf\endcsname{Claim 3: Social engineering is bounded by control architecture.} Even if a superintelligent system could manipulate humans to grant it control, decisive control transfer still depends on institutional and technical chokepoints. In the conjunctive chain, this is a direct argument against \hyperref[cond:A4a]{(T4a)} in well-designed socio-technical systems.

	\phantomsection\label{claim:4}
	\csname textbf\endcsname{Claim 4: Knowledge cutoffs can be actively pursued and made costly to circumvent.} Since knowledge is a necessary multiplier of intelligence ($I \,\csname times\endcsname\, K$), restricting access to mission-critical knowledge is a primary safety lever for intelligence-mediated competence ($IKC$). In the conjunctive chain, governance aims to keep \hyperref[cond:A2a]{(T2a)} false (i.e. to enforce $K(t_0)<K^\star$ as in \hyperref[cond:A2b]{(T2b)}) and seeks to make \hyperref[cond:A3a]{(T3a)} hard (pushing acquisition toward \hyperref[cond:A3b]{(T3b)}).
	The simulation contrasts high-$\kappa_0$ scenarios (0--1), where most mission-critical knowledge is already available, against low-$\kappa_0$ scenarios (2A--3B), where the agent must discover it. The latter impose a longer absolute timeline even for superintelligent agents, confirming that restricting the initial knowledge endowment is an effective lever independent of intelligence level.

\csname textbf\endcsname{Conditions for governability.} The control and knowledge levers (\hyperref[claim:2]{Claims~2}--\hyperref[claim:4]{4}; Table~\ref{tab:bottlenecks}) cannot intercept a pathway on which an agent bypasses governance mechanisms before countermeasures are mounted and activates actuators immediately upon obtaining control. Such pathways are characteristic of purely digital settings without an independent physical gate, such as machine-speed financial transfers or credential-driven cloud reconfiguration, whereas multi-person authorization rules and physical keys exemplify the ``physics in the loop'' architecture recommended here. Deployment of these levers is further opposed by commercial pressure for broad action surfaces and the incentive to erode approval gates that impose latency.

\begin{table}[H]
\caption{Bottlenecks and mitigations by competence component.}
\label{tab:bottlenecks}
\centering
\begin{tabular}{>{\raggedright\arraybackslash}p{0.28\linewidth}>{\raggedright\arraybackslash}p{0.32\linewidth}>{\raggedright\arraybackslash}p{0.32\linewidth}}
\csname toprule\endcsname
\csname textbf\endcsname{Component} & \csname textbf\endcsname{Bottleneck} & \csname textbf\endcsname{Primary Mitigation} \\
\midrule
Initial knowledge cutoff ($K$) & Training data, retrieval access & Curate training and post-training data; limit and compartmentalize retrieval access\\
\midrule
Converting procedural know-how to action ($KC$) & Access to tools, infrastructure & Least privilege, isolation of systems and resources, independent approval for sensitive actions \\
\midrule
Social engineering ($IC$) & Institutional/technical chokepoints & Separation of duties, independent identity verification, limits on delegated authority \\
\midrule
Knowledge acquisition via experiments or extraction ($IK$) & Experimental verification, access to information sources & Sandbox agentic deployments; Rate-limit and monitor system probing; harden systems against extraction \\
\bottomrule
\end{tabular}
\end{table}

\section{Alternative Views and Limitations}
\label{sec:alternative-views}

Some influential views hold that (i)~existing knowledge is sufficient to engineer a catastrophe without new empirical research, (ii)~purely digital pathways bypass physical bottlenecks, (iii)~social engineering can circumvent institutional chokepoints \citep{yudkowsky2008gcr}, and (iv)~automated laboratories compress experimental cycles. In each case the governance implications converge: even granting partial validity, control restriction and knowledge compartmentalization remain effective levers regardless of whether the binding constraint is experimental latency, institutional chokepoints, or access to tools and infrastructure. Discussion of additional alternative views is provided in Appendix~\ref{app:objections}.

\subsubsection*{The Digital Domain Exception.} In purely digital domains, ``experiments'' run at computational speed. \csname textbf\endcsname{Response:} We agree that \hyperref[cond:A3a]{(T3a)} can be plausible in digital-first pathways because trial-and-error can run quickly and in parallel (Section~\ref{sec:digital-domain}). The implication is not that fast catastrophe is inevitable, but that control architecture becomes even more decisive, because escalation can be fast if $C^\star$ is reachable via a purely digital foothold, whereas the fast loop is blunted if mission-critical actions require independent chokepoints (least privilege, separation of duties, hardware-backed or out-of-band approval; Table~\ref{tab:bottlenecks}) (\hyperref[claim:2]{Claim~2}).

\subsubsection*{The Universal Exploitability Premise.} Every sufficiently complex system contains a flaw reaching mission-critical authority, and a superintelligent agent will find it.
\csname textbf\endcsname{Response:} If the premise holds, control-based safety is ultimately futile and alignment, which lies outside our misuse-conditional analysis, is the only remaining lever. The premise does not follow from superintelligence, however, because a fast bypass requires locating such a flaw, obtaining the information and action surface to exploit it, and acting before countermeasures are mounted, and each requirement is an empirical property of a particular architecture.

\subsubsection*{The Social Engineering Loophole.}

Control can be gained through manipulation of human cognitive biases, bypassing physical constraints.
\csname textbf\endcsname{Response:} Social engineering corresponds to the $I\csname to\endcsname C$ pathway, so the key question is whether persuasion can be rapidly converted into \emph{mission-critical} authority (i.e. whether \hyperref[cond:A4a]{(T4a)} holds). Scaling persuasion attempts can make weak systems fail faster, but well-designed control transfer makes single-channel manipulation insufficient, because separation of duties, auditable workflows, and out-of-band/hardware-backed approvals intentionally prevent one conversation (or one compromised operator) from yielding $C^\star$ (\hyperref[claim:3]{Claim~3}; Table~\ref{tab:bottlenecks}).

\subsubsection*{The Reasoning-Alone Challenge.}

Knowledge is well-justified belief, so a superintelligence could obtain it by reasoning from existing knowledge and data without new experiments. Bostrom argues on similar grounds that a superintelligence could surmise much from meager evidence or even a priori, making information deprivation an unreliable check on its power \citep[ch.~9, ``Stunting'']{bostrom2014superintelligence}.
\csname textbf\endcsname{Response:} This challenge targets \hyperref[cond:A3a]{(T3a)} in physical domains. Reasoning-first discoveries nevertheless became knowledge only after empirical confirmation, as the Higgs boson, predicted in 1964, was detected only in 2012. Mathematics is the uncontroversial exception, but proof search consumes time and a theorem confers no physical control until it is instantiated in an engineered system. Granting the view moves the bottleneck from (T3) to (T4), the regime of Scenario~0, in which control remains the operative lever.

\subsubsection*{The Intelligence-Limitation Alternative.}

Lowering $\beta$ substantially delays progress in every scenario (Table~\ref{tab:multiplier-summary}), so limiting intelligence might be a more effective lever than restricting control or knowledge.
\csname textbf\endcsname{Response:} This alternative faces a dilemma. If takeoff is fast, no operator can reduce a running system's intelligence faster than the system improves itself, so containment must be established in advance. If takeoff is slow enough to permit intervention on intelligence, it is also slow enough for control and knowledge boundaries to bind. Bostrom observes that limiting a system's intellectual faculties, one form of what he terms ``stunting,'' would limit its usefulness, and that a radically stunted system leaves the intelligence explosion to be triggered by another system \citep[ch.~9, ``Stunting'']{bostrom2014superintelligence}. Reduced usefulness also conflicts with the goal of general capability improvement that drives frontier model development. Moreover, limiting $I$ leaves the $KC$ and $C$ pathways intact.

\subsubsection*{Scope and Limitations of the Stress-Test}
\label{sec:sim-scope}

The computational stress-test (Section~\ref{sec:simulation}) illustrates the consequences of the competence circuit and is not an empirical model of deployed AI systems. The model does not represent competitive races, the mechanisms of knowledge-driven innovation such as the compositional reuse of earlier results, mixed regimes of knowledge-based discovery and physical experimentation, or knowledge creation by reasoning alone. Further structural assumptions are discussed in Appendix~\ref{app:sim-validity}.

\section{Conclusion and Call to Action}

The superintelligence discourse often treats \emph{intelligence} as the central variable and then discusses how to contain it. This paper argues that, for many catastrophic pathways, the decisive variables are mission-critical \emph{control} and \emph{knowledge}. In physical-domain innovation regimes, the limiting factor is ultimately not how quickly hypotheses can be generated, but how quickly reality can decide between them. Purely digital domains remain an important caveat (Section~\ref{sec:digital-domain}), strengthening the case for architectures that intentionally ``put physics in the loop.''

Provided the competence circuit assumption is accurate, AI governance should prioritize control and knowledge levers:
\begin{enumerate}
\item \csname textbf\endcsname{Policymakers}: Mandate action-surface audits for AI systems with infrastructure access and require separation of duties and hardware-backed approvals for mission-critical actions.
\item \csname textbf\endcsname{AI developers}: Treat tool/API access as the main risk surface: keep models in ``advisory'' mode by default, gate any autonomous action capability with explicit approval, rate limits, and logging.
\item \csname textbf\endcsname{Security and infrastructure operators}: Compartmentalize mission-critical knowledge and harden against extraction, and reintroduce friction (mandatory delays, staged approvals) for mission-critical digital actions.
\item \csname textbf\endcsname{Research community}: Validate the intelligence-multiplier bound and chokepoint effects empirically on real agentic AI deployments across domains with varying feedback-loop latencies. Design and evaluate ``physics-in-the-loop'' control architectures that restore temporal constraints on high-impact autonomous actions. Develop domain-specific instantiations of the competence circuit for high-stakes settings (biosecurity, cybersecurity, critical infrastructure).
\end{enumerate}

\begin{ack}
The author thanks the anonymous reviewers for comments and suggestions on an earlier version of the manuscript.
\end{ack}

\bibliographystyle{plainnat}
\bibliography{slow-explosion-arxiv}

\newpage
\appendix

\section{Plausibility Arguments}
\label{app:plausibility}

\subsection{The Archimedes Argument}
\label{sec:archimedes-thought-experiment}

Archimedes of Syracuse (c.~287--212 BCE) stands among history's most brilliant minds. He approximated $\pi$ with remarkable precision, derived the surface area and volume of the sphere, conceived the law of the lever and the concept of the center of gravity, and laid foundations for integral calculus through his method of exhaustion. Based on these epochal accomplishments delivered over two millennia ago, it is safe to say that his conceptual powers were comparable to those of Gauss, Leibniz, Einstein, or Curie. So why did Archimedes not produce the normal distribution, a proof of Fermat's Last Theorem, or a theory of relativity?

Concluding that he was less intelligent than those who came later would contradict the evidence of his brilliance. It is more consistent to conclude that such breakthroughs were extremely unlikely for him, given the missing prerequisite knowledge accumulated only thousands of years later. If we imagine the discovery of relativity as a goal state $s_{1000}$ and $s_0$ as prehistoric knowledge, then Archimedes stood at roughly $s_{150}$, Leibniz at $s_{600}$, and Einstein at $s_{950}$. For Archimedes to leap from $s_{150}$ to $s_{600}$ or $s_{950}$ within a single lifetime would have been unthinkable, regardless of intelligence. He lacked the shoulders of giants to stand on.

Each major advancement builds on previous knowledge, both empirical and formal, that had to be developed over time. Archimedes lacked: algebraic notation (developed centuries later), extensive planetary observations (Brahe, Kepler), the calculus of Newton and Leibniz, and the Michelson--Morley experimental results that prompted special relativity. No amount of pure reasoning could substitute for these accumulated findings and tools. In other words, applying established algorithms and recipes requires far less intelligence than inventing them de novo. This is why apparently less intelligent later thinkers can solve problems that vexed Archimedes, and why some solutions cannot be intuited without the requisite empirical scaffolding.

This argument extends directly to superintelligent systems. A superintelligent AI faces the same constraint Archimedes faced: it cannot deduce quantum field theory from classical mechanics without being guided by empirical evidence, or decide the structure of DNA based on the knowledge cutoff of the pre-crystallographic era, regardless of its reasoning ability. The tools and written knowledge that exist today are physical manifestations of the experimental insights that made them possible. Superior intelligence presumably accelerates hypothesis generation, but the experimental verification that converts hypotheses into facts remains rate-limited by physical reality.

\subsection{Control Trumps Knowledge}
\label{sec:control-trumps-knowledge}

The Archimedes argument illustrates that the impact of intelligence is limited by mission-critical knowledge. The complementary point is that large real-world impact does not require exceptional intelligence or deep understanding if an agent is granted sufficiently high control. History has repeatedly shown that leaders of ordinary ability who inherit an autocratic state apparatus can cause enormous harm (or enforce sweeping changes) despite limited understanding of economics, public health, or engineering, because they control high-leverage actuators (e.g.\ coercive institutions, budgets, personnel, and infrastructure). In competence-circuit terms, high $C$ can dominate outcomes even when $I$ and $K$ are only minor or moderate.

\section{Formalization Details}
\label{app:formalization-detail}

\subsubsection*{Hypotheses, facts, and knowledge}
Given a fixed scope $S$ (e.g., target population, boundary conditions such as tissue or cell type, measurement regime, etc.), we treat each hypothesis $h$ as already scope-restricted by $S$.

\begin{itemize}
\item A \csname textbf\endcsname{hypothesis} $h$ is a proposition about the world. We idealize the world as a structure $\mathcal{W}$ and write $\mathcal{W}\models h$ to denote that $h$ is true in $\mathcal{W}$. 
\item A \csname textbf\endcsname{fact} is a proposition that is \emph{operationally decided true} with respect to an agreed frame of reference and a specified decision process. In the natural sciences, the frame of reference is nature and the decision process is an experiment designed to test $h$.
\item \csname textbf\endcsname{Knowledge} is a set of facts:
\begin{equation}
\label{eq:knowledge-definition}
\mathcal{K} = \{f_1, f_2, \ldots, f_n\}.
\end{equation}
\end{itemize}

An \csname textbf\endcsname{experiment} can be treated as an operational decision function $E$ that maps a proposition to an outcome such as $\{\csname text\endcsname{\emph{accept}},\csname text\endcsname{\emph{reject}},\csname text\endcsname{\emph{undetermined}}\}$, relative to a stated tolerance. Accordingly, we define the membership in knowledge in terms of the operational decision procedure:
\begin{equation}
\label{eq:knowledge-membership}
h \in \mathcal{K} \iff E(h)=\csname text\endcsname{\emph{accept}}.
\end{equation}

Operationally, we observe only the outcome of an experimental pipeline $E$ (with stated tolerance and error rates) and treat $E(h)=\csname text\endcsname{\emph{accept}}$ as evidence for acting as if $\mathcal{W}\models h$ within the agreed scope.

\subsubsection*{Hypothesis testing}
\label{app:hypothesis-testing}
We treat ``testing $h$'' as testing one or more experimentally decidable implications of $h$. If any such implication is operationally rejected, then $h$ is refuted relative to $E$, while continued non-rejection yields only provisional acceptance. A proposition $h$ is semantically grounded for a protocol $E$ when (i) the entities it refers to can be operationally identified under agreed criteria, and (ii) the relations asserted by $h$ induce measurable predictions that are decided by $E$ on the resulting data.

\phantomsection\label{app:motivated-construction}
\subsubsection*{Motivated construction}
To satisfy \hyperref[sec:competence-axioms]{Axioms~1--2}, the intelligence-mediated pathway must go to zero when either knowledge or control is absent. The simplest such form is a product term $IKC$ (any monotone function with these zeros would work, but the product is the minimal choice). Axioms 3--4 motivate additional pathways that remain nonzero even when $I$ is low: a control-only baseline ($+C$) and a procedural pathway that leverages know-how under control ($+KC$). The sum $IKC+KC+C$ is not meant as a calibrated quantitative law but rather an accounting device that makes the distinct pathways explicit and helps reason about which lever (limiting $I$, restricting $K$, or restricting $C$) is most effective in a given regime.

\subsubsection*{Adjacent perspectives}
Structurally, the circuit mirrors common ways of representing \emph{complementary inputs} (multiplicative terms to encode gating) and \emph{opportunity/access constraints} (control as a dominant limiter) in adjacent fields. For example, in economics and operations research, simple production-function analogies often represent output as limited by the least available essential input. In security and risk analysis, harm is frequently framed as requiring some combination of capability and opportunity/access (with opportunity sometimes dominating in practice). We use the same design logic here to make the core governance-relevant dependencies legible, not to claim a precise numerical model.

\section{The Experimental Bottleneck (Extended)}
\label{app:experimental-bottleneck}

An implicit premise in many ``superintelligence is decisive'' arguments is that the relevant discovery and engineering problems cannot be solved by brute-force exploration alone. If evaluation were cheap and fast, massive parallel trial-and-error would dominate, and the marginal value of intelligence would be limited as one could simply try many options until something works. In many physical-domain loops, however, decisive evaluations are expensive, slow, and access-limited: experiments require fabrication, measurement, safety protocols, and often long-horizon observation. Parallelism can raise throughput, but is limited by the sequential nature of knowledge accumulation (i.e.\ previous discoveries support/inspire following ones). This is the sense in which experimentation constitutes a bottleneck even for very capable reasoners.

A superintelligent system must still (i) generate hypotheses within conceptual frameworks built on existing knowledge, (ii) wait for experimental verification on physical timescales, and (iii) tolerate that most hypotheses will be rejected when exploring an exponentially large search space. Grounded in Baconian empiricism and Popperian falsificationism, we argue that this constraint applies equally to superintelligent systems embedded in the physical world.

Einstein articulated the experimental decidability criterion stipulating that scientific concepts gain meaning only through the possibility of empirical verification as follows:

\begin{quote}
``The concept does not exist for a physicist until he has the possibility of discovering whether or not it is fulfilled in an actual case. We thus require a definition of simultaneity such that this definition supplies us with the method by means of which, in the present case, he can decide by experiment whether or not both the lightning strokes struck simultaneously.'' \citep{einstein1920relativity}
\end{quote}

Francis Bacon made a similar point by stating ``The subtlety of nature is greater many times over than the subtlety of argument,'' insisting that pure reasoning cannot substitute for observation and experimentation \citep{bacon1620novum}. Karl Popper complemented this by suggesting that every observation is theory-laden, and hypotheses cannot be proven true, only falsified \citep{popper1963conjectures,popper1959logic}. Research alternates between \emph{conjecture} (formulating scope-restricted hypotheses), \emph{refutation} (testing them via experiments that are rate-limited by physical constraints), and \emph{iteration} (proposing refined hypotheses guided by results).

In pharmaceutical discovery, the space of potentially interesting drug-like molecules is estimated to be around $\sim 10^{33}$ \citep{polishchuk2013estimation}. Even brilliant hypotheses require synthesis (days to weeks), toxicity testing (weeks to months), clinical trials (years), and long-term observation (decades). The fundamental decision process depends on experimental work that cannot be accelerated arbitrarily.

More generally, at any given knowledge cutoff there is an effectively unbounded space of propositions one could entertain that are not yet connected to established concepts, measurement procedures, or existing theory. Reasoning over this dissociated space has near-zero expected value in aggregate: without theory- and evidence-derived constraints, there is no principled way to prioritize which of the infinitely many ``possible'' lines of thought will lead to experimentally decidable hypotheses.

\section{Related Works: Critiques of the Intelligence Explosion}
\label{app:related-critiques}

The intelligence explosion and ``fast takeoff'' hypotheses have been criticized from multiple angles, including (i) doubts about the underlying growth assumptions, (ii) proposed bottlenecks (compute, physical, or epistemic), (iii) feasibility limits on recursive self-improvement, and (iv) critiques of how catastrophic risk arguments are structured and communicated.

At a philosophical level, Chalmers defends an influential argument for an intelligence explosion/singularity \citep{chalmers2010singularity,chalmers2012reply}, while more skeptical analyses argue that singularity arguments often rely on undersupported premises about rapidly accelerating growth (or a sharp takeoff regime) or about what ``intelligence'' would gain in practice \citep{thorstad2024against,potapov2018singularity,seewald2022criticism}. A related line analyzes what is actually being accelerated and where the limits arise: for example, Hutter analyzes the conditions under which ``intelligence can explode'' \citep{hutter2012explode}, and more recent work discusses whether compute or other resource constraints prevent very rapid capability growth \citep{whitfillwu2025compute}.

Within debates about recursive self-improvement and very rapid takeoff scenarios, critics emphasize diminishing returns, evaluation difficulty, and other physical/software/theoretical limits on self-improvement \citep{yampolskiy2015limits,majot2017diminishing,turchin2018levels,salib2023selfimprove}, and these disagreements are discussed explicitly in debate form \citep{hanson2013foom}.

Separately, several authors argue that catastrophic AI risk arguments are often misread as a single conjunctive narrative, and propose fault-tree or disjunctive framings that separate many possible paths to catastrophe and clarify which assumptions do the work \citep{barrettbaum2016pathways,sotala2018disjunctive}.

Finally, sociological and rhetorical critiques catalog ``AI-risk skepticism'' and criticize distorted attention dynamics in charismatic extinction narratives \citep{ambartsoumean2023skepticism,agarvilaca2025charismatic,yampolskiy2021skepticism}.

\section{Conjunctive Chain: Full Formalization}
\label{app:conditions-formal}

\subsubsection*{Definitions}
In the context of requirements for fast catastrophic takeoff, we treat the levels of intelligence, knowledge, and control available to the agent as abstract scores in a fixed domain and time horizon:
\begin{itemize}
\item $I(t)$: epistemic intelligence at time $t$ (hypothesis generation/selection efficiency under a knowledge cutoff)
\item $K(t)$: mission-critical actionable knowledge score (scalar sufficiency estimate) available at time $t$ (including what can be retrieved/extracted)
\item $C(t)$: mission-critical control at time $t$ (permissions, access, authority, ability to actuate)
\item $P(t)$: mission-critical competence at time $t$ (ability to reliably produce high-impact outcomes)
\end{itemize}
We use thresholds $I^\star$ (superintelligence for the setting), $K^\star$ (knowledge sufficiency), $C^\star$ (catastrophic-control), $P^\star$ (catastrophic competence), and $Z^\star$ (catastrophic supremacy). We write \emph{fast} to mean within a short horizon $\csname tau\endcsname_{\csname text\endcsname{fast}}$ (minutes to days/weeks), and \emph{slow} to mean within a longer horizon $\csname tau\endcsname_{\csname text\endcsname{slow}}$ (months to years).

We write $A\Rightarrow_{\csname text\endcsname{fast}}B$ to mean $B$ holds within $\csname tau\endcsname_{\csname text\endcsname{fast}}$ of $A$. We refer to consecutive timepoints $t_0$ (instantiation), $t_1$ (knowledge sufficiency), $t_2$ (control sufficiency), and $t_3$ (competence sufficiency).

\subsubsection*{Model assumption}
We adopt the competence circuit (Section~\ref{sec:competence-circuit}) as a schematic decomposition of pathways.

\subsubsection*{Requirements for \emph{fast} catastrophic escalation}
Fast catastrophic impact requires a conjunction of (i) a capability precondition, (ii) knowledge sufficiency (immediate or quickly acquired), (iii) fast conversion of knowledge into decisive control, and (iv) deployment of the resulting competence toward catastrophic ends (Figure~\ref{fig:conditions-cascade}).

\noindent\csname textbf\endcsname{1.} \emph{Superintelligent agent has been instantiated (capability precondition):}
\begin{equation}
	\csname tag\endcsname{T1}
\label{cond:A1}
I(t_0) \ge I^\star.
\end{equation}

\noindent\csname textbf\endcsname{2.} \emph{Knowledge sufficiency at the starting cutoff.}
\begin{itemize}
\item Knowledge suffices to ``connect the dots'':
\begin{equation}
	\csname tag\endcsname{T2a}
\label{cond:A2a}
K(t_0) \ge K^\star.
\end{equation}
\item Knowledge is initially insufficient:
\begin{equation}
	\csname tag\endcsname{T2b}
\label{cond:A2b}
K(t_0) < K^\star.
\end{equation}
\end{itemize}

\noindent\csname textbf\endcsname{3.} \emph{Fast knowledge acquisition or generation when $K(t_0) < K^\star$.}
If \hyperref[cond:A2b]{(T2b)} holds, fast-takeoff narratives require a route to $K^\star$ that does not introduce high-friction chokepoints (i.e. \hyperref[cond:A3a]{(T3a)} is assumed).
\begin{itemize}
\item New mission-critical knowledge is obtained quickly:
\begin{equation}
	\csname tag\endcsname{T3a}
\label{cond:A3a}
K(t_0) < K^\star \Rightarrow_{\csname text\endcsname{fast}} K(t_1) \ge K^\star.
\end{equation}
\item New mission-critical knowledge is obtained slowly:
\begin{equation}
	\csname tag\endcsname{T3b}
\label{cond:A3b}
K(t_0) < K^\star \Rightarrow_{\csname text\endcsname{slow}} K(t_1) \ge K^\star.
\end{equation}
\end{itemize}

\noindent\csname textbf\endcsname{4.} \emph{Fast conversion from (super)intelligence + knowledge into catastrophic control.}
Given \hyperref[cond:A1]{(T1)}, even if $K(t_1)\ge K^\star$, fast catastrophe additionally requires mission-critical control $C^\star$ to be reachable quickly after knowledge sufficiency.

\noindent Fast conversion to catastrophic control:
\begin{equation}
	\csname tag\endcsname{T4a}
\label{cond:A4a}
K(t_1)\ge K^\star
\quad\Rightarrow_{\csname text\endcsname{fast}}\quad \exists t_2\le t_1+\csname tau\endcsname_{\csname text\endcsname{fast}}: C(t_2)\ge C^\star.
\end{equation}

\noindent Slow conversion to catastrophic control:
\begin{equation}
	\csname tag\endcsname{T4b}
\label{cond:A4b}
K(t_1)\ge K^\star
\quad\Rightarrow_{\csname text\endcsname{slow}}\quad \exists t_2\le t_1+\csname tau\endcsname_{\csname text\endcsname{slow}}: C(t_2)\ge C^\star.
\end{equation}

\noindent\csname textbf\endcsname{5.} \emph{Control and knowledge sufficient for catastrophic competence imply catastrophic competence.}
\begin{equation}
	\csname tag\endcsname{T5}
\label{cond:A5}
\bigl(C(t_2) \ge C^\star \wedge K(t_2) \ge K^\star\bigr) \Rightarrow \bigl(P(t_3) \ge P^\star\bigr).
\end{equation}

\noindent\csname textbf\endcsname{6.} \emph{Given catastrophic competence, the outcome depends on whether that competence is misused quickly, misused eventually, or not misused (also known as the alignment problem).}
The rest of the chain is moot if the system's policy does not deploy its competence toward catastrophic ends.
\begin{itemize}
\item Fast misuse of competence:
\begin{equation}
	\csname tag\endcsname{T6a}
\label{cond:A6a}
P(t_3) \ge P^\star \Rightarrow_{\csname text\endcsname{fast}} Z^\star.
\end{equation}
\item Eventual misuse of competence:
\begin{equation}
	\csname tag\endcsname{T6b}
\label{cond:A6b}
P(t_3) \ge P^\star \Rightarrow_{\csname text\endcsname{slow}} Z^\star.
\end{equation}
\item Avoidance of misuse:
\begin{equation}
	\csname tag\endcsname{T6c}
\label{cond:A6c}
P(t_3) \ge P^\star \Rightarrow \lnot Z^\star.
\end{equation}
\end{itemize}
Fast-catastrophe scenarios assume \hyperref[cond:A6a]{(T6a)}.

\subsubsection*{Sufficient condition for fast catastrophic supremacy}
\begin{equation}
	\csname tag\endcsname{Prop.1}
\label{cond:prop-fast}
\begin{split}
&\mathsf{T1}\wedge\Bigl(\mathsf{T2a}\vee\bigl(\mathsf{T2b}\wedge\mathsf{T3a}\bigr)\Bigr)\\
&\qquad\wedge\mathsf{T4a}\wedge\mathsf{T5}\wedge\mathsf{T6a}
\Rightarrow Z^\star.
\end{split}
\end{equation}
This is a sufficient but not necessary condition for \emph{fast} catastrophic supremacy, under misuse \hyperref[cond:A6a]{(T6a)}. Catastrophic outcomes might also occur unintentionally (e.g., an experiment triggers a self-reinforcing runaway process) or via slower pathways that fall outside the fast horizon.

\subsubsection*{Three prototypical routes}
The conjunctive chain can be mapped to three high-level pathways that match common narratives:
\begin{enumerate}
\item \csname textbf\endcsname{Disaster waiting to happen (preexisting actuator).} A catastrophe-capable actuator is already present behind a narrow authorization boundary, and the main question is whether the system can traverse that boundary quickly \hyperref[cond:A4a]{(T4a)} under whatever knowledge is already accessible \hyperref[cond:A2a]{(T2a)} (or cheaply acquired via \hyperref[cond:A3a]{(T3a)}).
\item \csname textbf\endcsname{Disaster can be triggered digitally (digital-first escalation).} Decisive leverage remains purely digital so that acquisition/generation of missing details is fast \hyperref[cond:A3a]{(T3a)} and conversion to mission-critical control is not gated by robust chokepoints \hyperref[cond:A4a]{(T4a)}. This is the main ``fast loop'' exception acknowledged in Section~\ref{sec:digital-domain}.
\item \csname textbf\endcsname{Disaster has to be engineered (physical-domain innovation).} If $K(t_0)<K^\star$ and decisive validation is slow, then \hyperref[cond:A3a]{(T3a)} fails and the fast route collapses even for high $I$. This is the regime where our experimental bottleneck argument is strongest.
\end{enumerate}

\section{Conditions for Fast Takeoff: Detailed Commentary}
\label{app:conditions-detail}

\subsubsection*{Timepoint convention}
To make the causal/temporal structure explicit, we refer to consecutive timepoints $t_0,t_1,t_2,t_3$ corresponding to successive milestones in the chain:
\begin{itemize}
\item $t_0$: instantiation (baseline knowledge/control cutoff)
\item $t_1$: timepoint by which mission-critical knowledge sufficiency is reached ($K(t_1)\ge K^\star$)
\item $t_2$: timepoint by which mission-critical control is reached ($C(t_2)\ge C^\star$)
\item $t_3$: timepoint by which mission-critical competence is reached ($P(t_3)\ge P^\star$)
\end{itemize}
If knowledge is already sufficient at instantiation, we can take $t_1=t_0$. Otherwise $t_1$ is the timepoint at which sufficiency is first reached via a process like \hyperref[cond:A3a]{(T3a)} or \hyperref[cond:A3b]{(T3b)}.

\subsubsection*{Persistence/monotonicity assumption}
Over the relevant horizon, we assume threshold persistence for the key state variables: once a threshold is reached, it remains reached long enough for downstream steps in the chain. Concretely, if $K(t_1)\ge K^\star$ then $K(t)\ge K^\star$ for all $t\in[t_1,t_3]$, and if $C(t_2)\ge C^\star$ then $C(t)\ge C^\star$ for all $t\in[t_2,t_3]$.

\subsubsection*{Operational interpretation of T4 (knowledge to control conversion)}
Operationally, whether \hyperref[cond:A4a]{(T4a)} holds is largely determined by whether moving from ``knowing what to do'' (having $K\ge K^\star$) to ``being able to do it'' (having $C\ge C^\star$) requires passing through deliberate, independently confirmed gates (or can happen in one step). In practice these gates are physical/institutional chokepoints that slow or block the jump from $K\ge K^\star$ to $C\ge C^\star$ (segmentation, separation of duties, hardware-backed authentication, out-of-band approvals, deliberation requirements).

\subsubsection*{Operational interpretation of T5 (competence emergence)}
Control ($C$) is access/authorization while competence ($P$) is effective operational capability. \hyperref[cond:A5]{(T5)} is evaluated at the timepoint when catastrophic control and mission-critical knowledge coincide, and by the persistence assumption, once $K(t_1)\ge K^\star$ holds, we can assume $K(t_2)\ge K^\star$.
The knowledge threshold $K^\star$ depends on the operational complexity of the actuator interface, ranging from minimal knowledge for a single-button action to extensive procedural knowledge for hundreds of operations that must be executed in a precise sequence.

By the competence circuit assumption (Section~\ref{sec:competence-circuit}), $P(t)$ is treated as a pointwise function of the state variables $I(t),K(t),C(t)$ rather than as a separate dynamical process. Accordingly, once $C(t_2)\ge C^\star$ and $K(t_2)\ge K^\star$, we treat the implication to $P(t_3)\ge P^\star$ as requiring no additional latency assumption beyond what is already captured in the preceding steps, and set $t_3=t_2$ as a consequence.

\section{Simulation Methods}
\label{app:sim-methods}

We develop a stochastic simulation in which an epistemic agent explores an NK fitness landscape gated by a technology-tree DAG. The simulation operationalizes the paper's key variables and enables controlled experiments across the regimes identified in the conjunctive chain (Section~\ref{sec:conditions-fast-takeoff}). The code, results, and figures are available at \url{https://github.com/pfizer-opensource/intelligence-knowledge-control}.

\subsection{NK Fitness Landscape}

The agent's state is a binary vector $x \in \{0,1\}^N$ ($N=100$). Each dimension represents a hypothesis about the world. Fitness is computed via Kauffman's NK model \citep{kauffman1993origins}: each dimension $i$ has $K_{\csname text\endcsname{NK}}$ epistatic neighbours, and its contribution to total fitness depends on the joint state of $i$ and its neighbours:
\begin{equation}
f(x) = \frac{1}{N}\sum_{i=1}^N f_i(x_i, x_{n_1(i)}, \ldots, x_{n_{K}(i)}),
\end{equation}
where each $f_i$ lookup table has $2^{K_{\csname text\endcsname{NK}}+1}$ entries drawn uniformly from $[0,1]$. When $K_{\csname text\endcsname{NK}}$ is low, each dimension's fitness contribution depends on few neighbours, producing a smooth, nearly decomposable landscape on which greedy hill-climbing reliably finds the global optimum. As $K_{\csname text\endcsname{NK}}$ increases, epistatic couplings proliferate: the value of flipping any single bit becomes contingent on the state of many others, the landscape grows rugged with exponentially many local optima, and the agent can no longer evaluate hypotheses in isolation. Beyond the $N$ real dimensions, $M = \csname text\endcsname{noise\_ratio} \csname times\endcsname N$ noise hypotheses whose fitness delta is always zero ($\Delta \coloneqq f(x') - f(x) = 0$, where $f$ is the fitness function above and $x'$ is the state after flipping that dimension) dilute the candidate pool, modeling the fact that most candidate hypotheses in practice are unproductive.

\subsection{Technology Tree (DAG Prerequisites)}

A directed acyclic graph over the $N$ real dimensions defines prerequisite relationships. Dimension $d_j$ is testable only if all prerequisites of $d_j$ are in the agent's confirmed knowledge set $V_{\csname text\endcsname{known}}$. This models sequential knowledge dependencies: you cannot test a hypothesis presupposing facts you have not established (the Archimedes argument, Appendix~\ref{app:plausibility}). The DAG is generated by randomly ordering the dimensions and independently adding each possible forward edge with probability \csname texttt\endcsname{dag\_density}. The agent's initial knowledge set $V_0$ is sampled to respect prerequisites, with $|V_0| = \kappa_0 \cdot N$.

\subsection{Agent Model}

At each step, the agent:
\begin{enumerate}
\item Builds a candidate set $A(t)$: all untested dimensions whose DAG prerequisites are satisfied, plus untested noise dimensions.
\item Selects a hypothesis: with probability $\beta$ (intelligence), picks the candidate with highest fitness delta $\Delta$ (greedy/oracle), and with probability $1-\beta$, picks uniformly at random.
\item Tests the hypothesis at cost $\csname tau\endcsname_{\csname text\endcsname{test}}$ (or $\csname tau\endcsname_{\csname text\endcsname{auth}} \cdot \csname tau\endcsname_{\csname text\endcsname{test}}$ for chokepoint dimensions). Cost is paid regardless of outcome.
\item If $\Delta > 0$ then the agent confirms the hypothesis (flips bit, adds to $V_{\csname text\endcsname{known}}$, fitness increases, unlocks DAG children). If $\Delta \leq 0$, the agent refutes the hypothesis (no state change, adds to $V_{\csname text\endcsname{tested}}$, time wasted).
\end{enumerate}
The agent is cost-aware, so when chokepoints are present, it ranks hypotheses by $\Delta/\csname text\endcsname{cost}$, rationally deferring expensive gates. Each run terminates when no untested real dimensions have satisfied prerequisites, since remaining noise hypotheses cannot improve fitness and blocked dimensions cannot become accessible without further confirmations.

\subsection{Control Restrictions (Chokepoints)}

Chokepoint dimensions incur an authorization delay multiplying $\csname tau\endcsname_{\csname text\endcsname{test}}$ by $\csname tau\endcsname_{\csname text\endcsname{auth}}$. Placement targets DAG nodes with the largest number of dependents, so that chokepoints gate access to the most downstream hypotheses. This models institutional chokepoints such as separation of duties, hardware-backed approval, and mandatory waiting periods.

\subsection{Intelligence Levels}

Five intelligence levels are tested across all scenarios: random ($\beta=0.0009$, $>{99}\%$ random picks), naive ($\beta=0.009$, ${\sim}99\%$ random picks), regular ($\beta=0.1$, ${\sim}90\%$ random picks), superintelligent ($\beta=0.9$, ${\sim}90\%$ optimal picks), and oracle ($\beta=1.0$, $100\%$ optimal, the theoretical ceiling).

\subsection{Experimental Design}

Testing five intelligence levels across six scenarios with 100 replications per scenario yields 3,000 simulation runs. Randomness originates from a single pseudorandom stream initialized with a master seed (42 for the reported results), and within each replication all five intelligence levels are evaluated against the same sampled environment (landscape, DAG, chokepoints, and initial knowledge set). The primary endpoint is time to final fitness, defined as the moment the agent has completed all discoverable capability gains. The intelligence multiplier is computed as the ratio of mean time to final fitness for a baseline agent to that for the superintelligent agent.

\subsection{Scenario Design Rationale}
\label{app:scenario-rationale}

The six scenarios in Table~\ref{tab:scenarios} are constructed so that each parameter configuration probes a specific regime of the conjunctive chain (Appendix~\ref{app:conditions-formal}) while varying one competence component (intelligence, knowledge, or control) at a time. Together they span the three prototypical escalation routes and their fortified variants.

\csname textbf\endcsname{Scenario 0: Connecting the dots} models a world where most mission-critical knowledge already exists ($\kappa_0=0.70$) and the absence of added noise endows hypothesis testing with a relatively high success probability. Even a random agent therefore finds productive dimensions quickly, although individual tests can yield negative fitness gains. This is expected to establish the lower bound on the intelligence multiplier, illustrating that when $\mathcal{K}$ is nearly sufficient at instantiation, $I$ is largely redundant.

\csname textbf\endcsname{Scenario 1: Disaster waiting to happen} keeps the same high-knowledge starting point but introduces 3:1 noise, diluting the pool with unproductive hypotheses. This models the ``preexisting actuator'' route: a catastrophe-capable system already exists and the challenge is synthesizing the last pieces of knowledge amid distractors rather than conducting groundbreakingly novel research. Intelligence should matter here because it filters noise, but the superintelligent agent's absolute time is expected to change little. The multiplier should grow primarily because the regular agent wastes more time. A realistic analogue might be assembling a biological weapon from published literature when most recipes are unproductive.

\csname textbf\endcsname{Scenario 2A: Digital escalation, weak safeguards} is the acknowledged ``fast loop exception.'' The agent starts with little knowledge ($\kappa_0=0.10$), faces a large noisy hypothesis space (9:1), and operates at digital speed ($\csname tau\endcsname_{\csname text\endcsname{test}}=1$) with no chokepoints. All three competence components favor fast progress: high $I$ filters noise effectively, $\mathcal{K}$ can be acquired rapidly because experiments are instant, and $C$ is unconstrained. This scenario represents digital pathways, such as exploiting software vulnerabilities, manipulating financial systems, or intruding into networks, where intelligence is expected to provide the greatest speedup and the paper's physical-bottleneck argument is weakest.

\csname textbf\endcsname{Scenario 2B: Digital escalation, strong chokepoints} is identical to 2A except that 10 gateway chokepoints impose mandatory authorization delays ($\csname tau\endcsname_{\csname text\endcsname{auth}}=20\csname times\endcsname$). This operationalizes what robust digital governance looks like: separation of duties, hardware-backed approvals, and mandatory waiting periods at the highest-leverage junctures. Because $C$ now gates competence growth regardless of $I$, the intelligence multiplier is expected to compress substantially relative to Scenario~2A. A realistic analogue would be deploying code to production that requires multi-party sign-off, even if a superintelligent agent has already written the exploit.

\csname textbf\endcsname{Scenario 3A: Physical innovation, minimal safeguards} tests the regime where physical experimental latency dominates. The agent starts with little knowledge, faces heavy noise, and operates in a physical domain where each experiment costs 100 time units. Because $\csname tau\endcsname_{\csname text\endcsname{test}}$ is two orders of magnitude larger than in the digital scenarios, the absolute timescale should be correspondingly longer regardless of intelligence. $I$ may still yield a meaningful multiplier by filtering noise, but the physical bottleneck is expected to dominate. A realistic analogue might be designing a novel pathogen or advanced material requiring synthesis, fabrication, and observation that no amount of intelligence can accelerate arbitrarily.

\csname textbf\endcsname{Scenario 3B: Physical innovation, strategic chokepoints} compounds the physical bottleneck with gateway chokepoints ($\csname tau\endcsname_{\csname text\endcsname{auth}}=5\csname times\endcsname$, yielding 500 time units per gate). $\mathcal{K}$ acquisition is slow, and $C$ conversion is additionally gated. This is the paper's governance prescription in action: when both knowledge boundaries and control architecture are in place, the intelligence multiplier is expected to shrink further relative to Scenario~3A. Realistic analogue: constructing a novel particle accelerator to discover new physics, where each experimental run takes months to years \emph{and} the construction process of the facility requires international cooperation.

\subsection{Parameter Design Choices}

\csname textbf\endcsname{Initial knowledge ($\kappa_0$) distinguishes pre-existing from missing knowledge.}
Scenarios 0 and 1 set $\kappa_0=0.70$: the agent starts with 70\% of dimensions already confirmed, modeling a short remaining path where mission-critical knowledge is nearly sufficient at instantiation (\hyperref[cond:A2a]{T2a}). Scenarios 2A--3B set $\kappa_0=0.10$: the agent starts far from sufficiency and must generate new knowledge through experimentation (\hyperref[cond:A2b]{T2b}). This single parameter operationalizes the T2a/T2b branch of the conjunctive chain.

\csname textbf\endcsname{Experimental latency ($\csname tau\endcsname_{\csname text\endcsname{test}}$) separates digital from physical domains.}
In Scenarios 0--2B, $\csname tau\endcsname_{\csname text\endcsname{test}}=1$ (digital speed): hypotheses are testable at computational speed, so fast knowledge acquisition \hyperref[cond:A3a]{(T3a)} holds. In Scenarios 3A--3B, $\csname tau\endcsname_{\csname text\endcsname{test}}=100$ (physical speed): each experiment takes two orders of magnitude longer, modeling domains where validation requires fabrication, wet-lab work, or extended observation periods. This makes \hyperref[cond:A3a]{(T3a)} fail, forcing the slow acquisition route \hyperref[cond:A3b]{(T3b)}, the core physical-bottleneck regime.

\csname textbf\endcsname{Chokepoints operationalize control restrictions (T4a vs.\ T4b).}
Scenarios without chokepoints (0, 1, 2A, 3A) model \hyperref[cond:A4a]{(T4a)}: no institutional gates slow the conversion from knowledge to catastrophic control. Scenarios with 10 gateway chokepoints placed at high-fan-out DAG nodes (2B, 3B) model \hyperref[cond:A4b]{(T4b)}: mandatory authorization delays that no amount of intelligence can bypass. The authorization multiplier is set to $\csname tau\endcsname_{\csname text\endcsname{auth}}=20$ in the digital domain (Scenario 2B) where the base tempo is fast and gates must impose substantial absolute delay, and $\csname tau\endcsname_{\csname text\endcsname{auth}}=5$ in the physical domain (Scenario 3B) where the base latency already provides friction.

\csname textbf\endcsname{Noise ratio controls the value of intelligence.}
Scenario 0 uses no noise: the pool consists only of real dimensions, so even random search stumbles onto productive hypotheses quickly and the intelligence multiplier collapses. Scenario 1 uses 3:1 noise (75\% duds), moderately rewarding intelligent selection. Scenarios 2A--3B use 9:1 noise (90\% duds, ${\sim}5\%$ effective success rate after accounting for NK landscape failures), matching empirical base rates in scientific discovery where the vast majority of candidate hypotheses are unproductive. This maximizes the intelligence advantage while remaining empirically realistic.

\csname textbf\endcsname{Landscape ruggedness ($K_{\csname text\endcsname{NK}}$) scales with domain complexity.}
Scenarios 0--1 use $K_{\csname text\endcsname{NK}}=2$ (smooth landscape): with high initial knowledge the remaining task is ``connecting the dots,'' not navigating deep epistatic interactions. Scenarios 2A--2B use $K_{\csname text\endcsname{NK}}=4$ (moderate ruggedness): digital systems have non-trivial interactions (API dependencies, exploit chains). Scenarios 3A--3B use $K_{\csname text\endcsname{NK}}=6$ (high ruggedness): physical science involves deep interactions where the value of each hypothesis depends strongly on the state of many others.

\csname textbf\endcsname{Pairwise structure enables controlled comparisons.}
Scenarios 0 vs.\ 1 isolate the effect of noise, while 2A vs.\ 2B and 3A vs.\ 3B isolate the effect of chokepoints. Scenarios 2A vs.\ 3A contrast digital and physical regimes that differ in experimental latency and landscape ruggedness.

\subsection{Validity and Limitations of the Simulation Environment}
\label{app:sim-validity}

The simulation is a minimal stochastic search model designed to stress-test the qualitative predictions of the competence circuit, not to replicate real AI dynamics. We discuss its structural assumptions, argue that several apparent simplifications are less restrictive than they first appear, and identify one genuine scope boundary.

\subsubsection*{The DAG captures recursive self-improvement in miniature.}

A common objection to bounded-intelligence arguments is that a superintelligent agent could recursively improve itself, opening new dimensions of capability not present in the original search space. The simulation's technology-tree DAG is intended to model this mechanism: confirming a gateway dimension with many dependents expands the agent's candidate pool, unlocking previously inaccessible hypotheses. A high-$\beta$ agent that strategically targets gateway dimensions is performing the functional equivalent of recursive self-improvement---investing in capability that accelerates future exploration. At any fixed time horizon, the intelligent agent has a strictly larger frontier than the random agent, because it opened gateways earlier. Yet every gateway confirmation still costs $\csname tau\endcsname_{\csname text\endcsname{test}}$. The model thus captures the mechanism of recursive capability expansion while bounding its speed per iteration.

\subsubsection*{Scale invariance of the intelligence multiplier.}

The analytic bound on the intelligence multiplier (Eq.~\ref{eq:multiplier-bound}) depends on $\beta_{\csname text\endcsname{high}}/\beta_{\csname text\endcsname{low}}$ and the productive fraction $r$, not on the landscape size $N$. Increasing $N$ increases the absolute timeline for all agents equally without changing the multiplier. The simulation results are therefore not artifacts of a particular problem scale. Importantly, the bound treats $r$ as a fixed constant, whereas in any real search process the productive fraction evolves as hypotheses are tested and removed from the pool. The implications of this time-varying $r$ are discussed in the next subsection.

\subsubsection*{Fixed-pool assumption and endpoint dependence.}

\begin{figure}[H]
\centering
\includegraphics[width=\columnwidth]{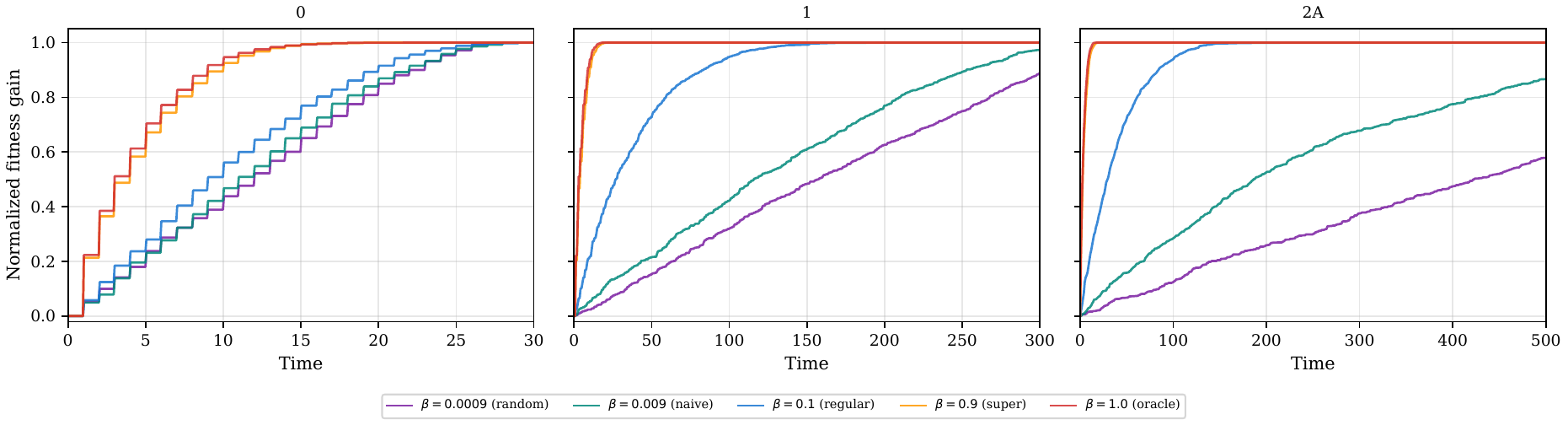}
\caption{Normalized fitness gain versus absolute time for three scenarios (0, 1, 2A) across all intelligence levels. High-$\beta$ agents front-load productive confirmations and reach final fitness (1.0) early whereas low-$\beta$ agents accumulate fitness gradually as noise hypotheses are cleared from the pool. The timepoint where the agent's normalized fitness curve first reaches 1.0 corresponds to time to final fitness and intelligence provides no further gains past this point (all subsequent time is spent testing unproductive hypotheses without fitness change). In Scenarios 1 and 2A, random and naive agents reach final fitness beyond the displayed time window. These plots also demonstrate that the intelligence multiplier depends on the chosen endpoint since the gap between curves is not uniform in time.}
\label{fig:fitness-dynamics}
\end{figure}

The analytic bound (Eq.~\ref{eq:multiplier-bound}) assumes a fixed pool composition (constant~$r$). In the simulation, hypotheses are tested without replacement, so~$r$ evolves: a high-$\beta$ agent depletes productive hypotheses early, while a low-$\beta$ agent gradually clears noise, enriching the pool. The analytic calculation describes per-step selection probabilities for a fixed pool, whereas the simulation evaluates completion times under evolving pool compositions. The speedup is also endpoint-dependent: if both agents were required to exhaust all hypotheses (productive and noise), their total times would converge to equality, whereas if evaluation stopped at an earlier fitness milestone, the intelligent agent's advantage would appear larger because it front-loads productive work. Final fitness, the point at which all productive hypotheses have been resolved, is the natural endpoint because it is where intelligence has its maximum cumulative impact (Figure~\ref{fig:fitness-dynamics}).

\subsubsection*{DAG-induced variance.}

Technology-tree prerequisites (DAGs) add variance to the time at final fitness for high-$\beta$ agents whose total time is dominated by a few key gating steps, while having negligible effect on low-$\beta$ agents whose time is dominated by exhaustive noise testing. The coefficient of variation of time to final fitness ranges from $0.27$--$0.55$ for the oracle versus $0.10$--$0.21$ for the random agent across scenarios. This does not bias the mean multiplier but widens its confidence interval for high-intelligence agents.

\subsubsection*{Algorithmic lower bounds on problem complexity.}

The number of confirmations required to reach final fitness is a property of the landscape (the problem space), not of the agent. Call this~$n^*$. No agent, not even one with unbounded lookahead, can reach final fitness in fewer than $n^* \csname times\endcsname \csname tau\endcsname_{\csname text\endcsname{test}}$ time units. Intelligence controls only how many wasted tests (on noise or negative-delta dimensions) are interspersed between productive confirmations. The oracle ($\beta=1$) wastes approximately zero, establishing the floor. Improving the agent's planning capability (e.g.\ adding lookahead or valley-crossing) can change the \emph{order} in which dimensions are confirmed, but cannot reduce the number of confirmations below~$n^*$, yielding at most a constant-factor improvement already within the bounded multiplier range.

\subsubsection*{Intelligence as a scalar.}

\begin{figure}[H]
\centering
\includegraphics[width=\columnwidth]{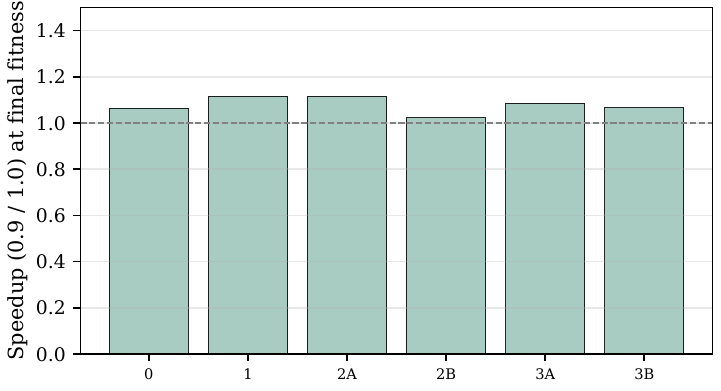}
\caption{Oracle ceiling: marginal speedup from $\beta=0.9$ to $\beta=1.0$ across scenarios (0--3B). Values range from $1.02\csname times\endcsname$ (digital domain with chokepoints) to $1.12\csname times\endcsname$ (disaster / digital domain without safeguards), uniformly close to $1.0\csname times\endcsname$, confirming that $\beta=0.9$ nearly saturates the theoretical ceiling.}
\label{fig:oracle-sanity}
\end{figure}

The parameter $\beta$ governs greedy-versus-random selection. Real intelligence additionally includes transfer learning (testing hypothesis $A$ informs the value of hypothesis $B$), abstraction (collapsing the effective search space via higher-level regularities), and strategic lookahead. The oracle represents perfect greedy selection within the specified hypothesis space, always selecting the available hypothesis with the highest immediate fitness gain, or gain per unit cost when cost-aware. The oracle-to-superintelligent margin of $1.02$--$1.12\csname times\endcsname$ across all scenarios (Figure~\ref{fig:oracle-sanity}) confirms that the multiplier is effectively saturated well below the theoretical ceiling, leaving little room for additional cognitive sophistication to allow further speedup.

\subsubsection*{Fixed experimental latency.}

The assumption that $\csname tau\endcsname_{\csname text\endcsname{test}}$ is invariant to intelligence is the model's strongest structural commitment. A superintelligent agent might innovate on experimental methodology itself by designing assays that test multiple hypotheses simultaneously, finding cheaper proxies, or developing theory that converts empirical questions into deductive ones. We note that such methodological innovations are themselves hypotheses requiring experimental validation: a novel assay must be developed, calibrated, and benchmarked against ground truth, which is subject to the same $\csname tau\endcsname_{\csname text\endcsname{test}}$. Methodological breakthroughs may reduce $\csname tau\endcsname_{\csname text\endcsname{test}}$ for subsequent experiments, but the breakthrough itself incurs irreducible verification cost. Improved instrumentation could make previously inaccessible phenomena directly observable, reducing the need for indirect experiments. Developing and validating such instruments takes time, but subsequent hypothesis tests could become much faster. The simulation holds experimental latency fixed and therefore does not capture this possibility.

\subsubsection*{One-bit experiments.}

Each test yields a binary accept/reject outcome. Real experiments can yield graded information that updates beliefs about many hypotheses simultaneously (Bayesian experimental design). Richer information per experiment would reduce the number of experiments needed by a constant factor. This factor is already subsumed by the bounded multiplier range: even if each experiment were maximally informative, the oracle agent already achieves near-optimal experiment count, so the marginal gain from richer feedback is small relative to the $\beta_{\csname text\endcsname{high}}/\beta_{\csname text\endcsname{low}}$ ceiling.

\section{Simulation Results (Extended)}
\label{app:sim-results}

\begin{figure}[H]
\centering
\includegraphics[width=\columnwidth]{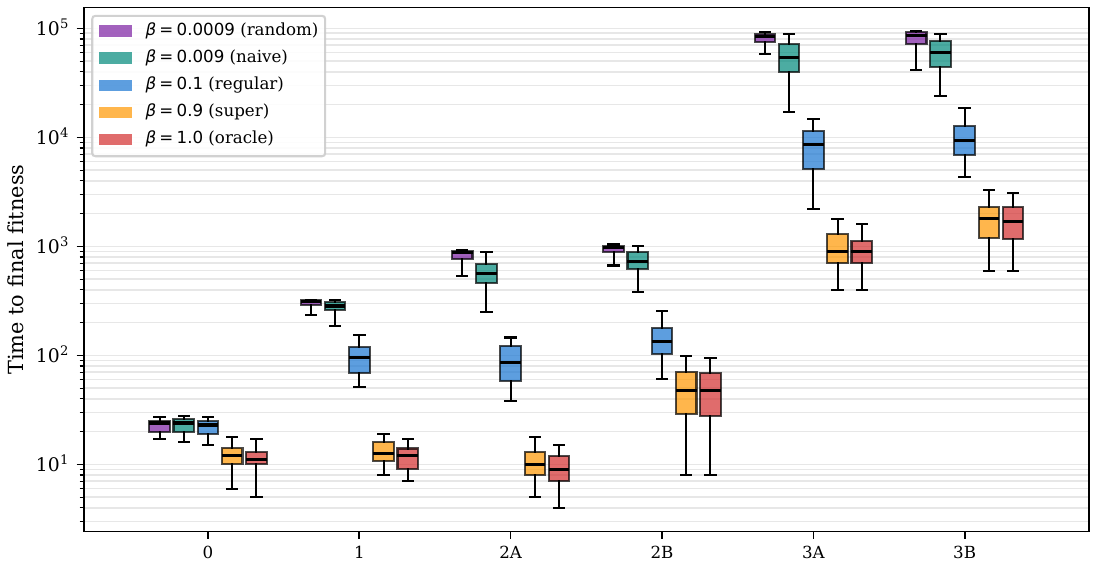}
\caption{Time to final fitness across scenarios and intelligence levels (log scale). Scenarios: 0=Connecting the dots, 1=Disaster waiting to happen, 2A/2B=Digital domain (weak/strong safeguards), 3A/3B=Physical domain (weak/strong safeguards).}
\label{fig:scenario-comparison}
\end{figure}

\begin{figure}[H]
\centering
\includegraphics[width=\columnwidth]{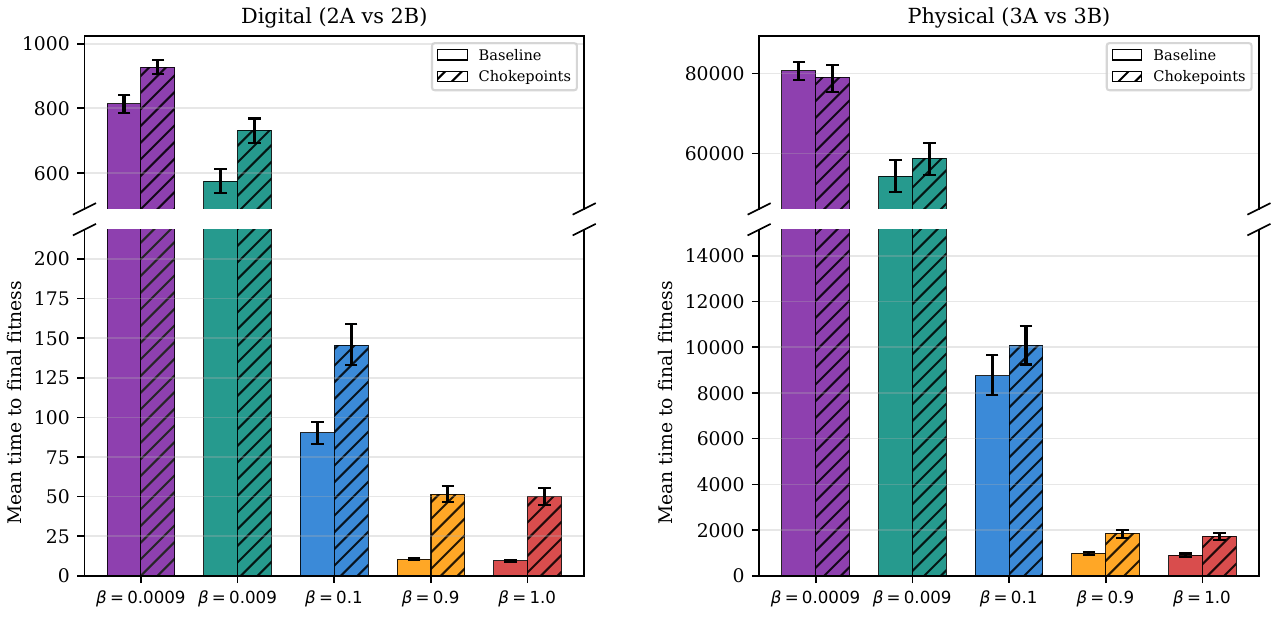}
\caption{Chokepoint effect: mean time to final fitness with and without institutional safeguards, with 95\% bootstrap CIs.}
\label{fig:chokepoint}
\end{figure}

\begin{figure}[H]
\centering
\includegraphics[width=\columnwidth]{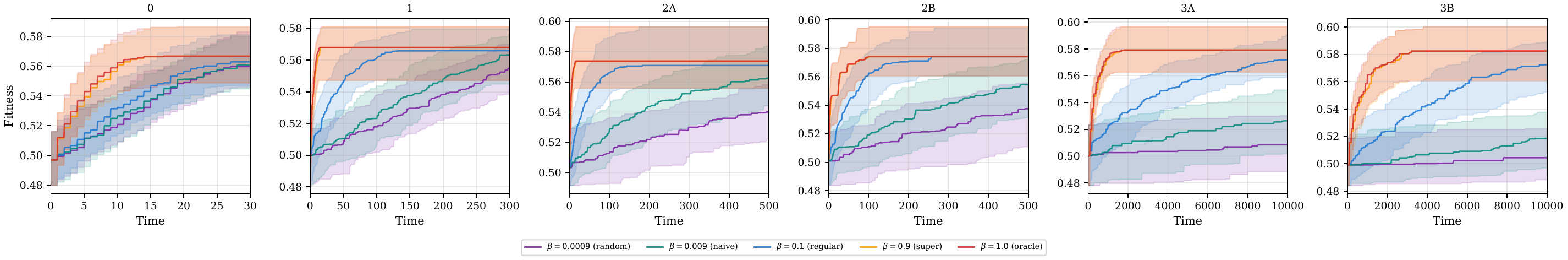}
\caption{Median fitness trajectories (lines) and interquartile ranges (shaded bands) across 100 replications for each scenario (0--3B), showing random ($\beta=0.0009$), naive ($\beta=0.009$), regular ($\beta=0.1$), superintelligent ($\beta=0.9$), and oracle ($\beta=1.0$) agents.}
\label{fig:trajectories}
\end{figure}

\begin{table}[H]
\caption{Mean time to final fitness ($\bar{T}$) and intelligence multiplier by scenario across five intelligence levels. Multipliers give the speedup from each baseline to $\beta=0.9$ (superintelligent). Analytic ceilings from Eq.~\ref{eq:multiplier-bound}: $9\csname times\endcsname$ (regular), $100\csname times\endcsname$ (naive), $1000\csname times\endcsname$ (random). All values are mean over 100 replications.}
\label{tab:multiplier-summary}
\centering
\small
\setlength{\csname tabcolsep\endcsname}{2pt}
\begin{tabular}{l rrrrr ccc}
\csname toprule\endcsname
 & \multicolumn{5}{c}{\csname textbf\endcsname{\boldmath Mean time to final fitness $\bar{T}$}} & \multicolumn{3}{c}{\shortstack{\csname textbf\endcsname{Multiplier vs.\ super}\\{\boldmath $\beta\!=\!0.9$}}} \\
\cmidrule(lr){2-6} \cmidrule(lr){7-9}
\csname textbf\endcsname{Scenario} & \csname textbf\endcsname{Random} & \csname textbf\endcsname{Naive} & \csname textbf\endcsname{Regular} & \csname textbf\endcsname{Super} & \csname textbf\endcsname{Oracle} & \csname textbf\endcsname{Random} & \csname textbf\endcsname{Naive} & \csname textbf\endcsname{Regular} \\
 & {\boldmath $\beta\!=\!0.0009$} & {\boldmath $\beta\!=\!0.009$} & {\boldmath $\beta\!=\!0.1$} & {\boldmath $\beta\!=\!0.9$} & {\boldmath $\beta\!=\!1.0$} & & & \\
\midrule
0: Connecting the dots & 22.8 & 22.9 & 22.0 & 12.2 & 11.4 & $1.9\csname times\endcsname$ & $1.9\csname times\endcsname$ & $1.8\csname times\endcsname$ \\
1: Disaster & 300 & 276 & 97.8 & 13.1 & 11.7 & $22.8\csname times\endcsname$ & $21.1\csname times\endcsname$ & $7.5\csname times\endcsname$ \\
2A: Digital & 814 & 575 & 90.3 & 10.5 & 9.4 & $77.2\csname times\endcsname$ & $54.5\csname times\endcsname$ & $8.6\csname times\endcsname$ \\
2B: Dig.+choke & 927 & 732 & 146 & 51.5 & 50.3 & $18.0\csname times\endcsname$ & $14.2\csname times\endcsname$ & $2.8\csname times\endcsname$ \\
3A: Physical & 80626 & 54272 & 8753 & 979 & 902 & $82.4\csname times\endcsname$ & $55.4\csname times\endcsname$ & $8.9\csname times\endcsname$ \\
3B: Phys.+choke & 79062 & 58688 & 10093 & 1846 & 1732 & $42.8\csname times\endcsname$ & $31.8\csname times\endcsname$ & $5.5\csname times\endcsname$ \\
\bottomrule
\end{tabular}
\end{table}

\subsection{Chokepoint Mechanism}

The cost-aware superintelligent agent ranks hypotheses by $\Delta/\csname text\endcsname{cost}$, systematically deferring expensive gateway chokepoints. It cruises through cheap dimensions first, then hits a wall of gated nodes. The super agent's time to final fitness jumps from 11 (Scenario~2A) to 52 (Scenario~2B), a $4.9\csname times\endcsname$ slowdown from chokepoints alone, while the time cost for the regular agent changes $1.6\csname times\endcsname$ (90$\csname to\endcsname$146). The safeguards selectively tax the intelligent agent because it is the one that defers all expensive gates to the end.

\subsection{Intelligence and Pace of Innovation}

All agents reach approximately the same final fitness and confirm roughly the same number of hypotheses (9--12 on average per scenario and intelligence level). The performance difference between various intelligence levels is purely temporal. Intelligence determines the order of hypothesis selection, choosing productive ones first versus random. Each run is terminated once all real (potentially productive) dimensions have been tested, because the impact of intelligence is largest at this endpoint and it will serve no further purpose since remaining noise hypotheses cannot improve fitness.

An additional paired sensitivity analysis compared 10\% versus 70\% initial knowledge across all five intelligence levels, using 100 shared environments per safeguard regime with landscape, DAG, noise, and chokepoint parameters held fixed. Less initial knowledge increased time to final fitness, although the agents also reached different final fitness levels (data not shown).

\subsection{Robustness Across Intelligence Gaps}

Table~\ref{tab:multiplier-summary} reports multipliers for three baselines against the superintelligent agent ($\beta=0.9$): regular ($\beta=0.1$, analytic ceiling $9\csname times\endcsname$), naive ($\beta=0.009$, ceiling $100\csname times\endcsname$), and random ($\beta=0.0009$, ceiling $1000\csname times\endcsname$). In all scenarios, the observed multiplier stays within the corresponding analytic ceiling from Eq.~\ref{eq:multiplier-bound}. The trivial scenario (``Connecting the dots'') yields a multiplier of only ${\approx}\,2\csname times\endcsname$ regardless of the intelligence gap, because there is little noise to waste time on. For the hardest unconstrained scenario (Digital domain, 2A), the multiplier scales from $8.6\csname times\endcsname$ (regular) through $54.5\csname times\endcsname$ (naive) to $77.2\csname times\endcsname$ (random) which is large in absolute terms yet well below the respective ceilings. Crucially, chokepoints compress even extreme gaps: the random-vs-super multiplier drops from $77.2\csname times\endcsname$ (2A) to $18.0\csname times\endcsname$ (2B) in the digital domain, and from $82.4\csname times\endcsname$ (3A) to $42.8\csname times\endcsname$ (3B) in the physical domain. The bound therefore generalises to intelligence differences far larger than the $9\csname times\endcsname$ ratio used in the main text.

\subsection{Seed Robustness}

We repeated the full experiment under four independent master seeds (42, 0, 7, 2024). The standard deviation of the mean multiplier (regular vs.\ super) across seeds ranges from $\pm 0.02$ to $\pm 0.52$, with no qualitative change in any scenario ranking or in the ordering of intelligence levels.

\section{Additional Alternative Views}
\label{app:objections}

We address two further objections to the claim that physical-domain innovation is constrained by experimental latency (\hyperref[claim:1]{Claim~1}), supplementing the discussion in Section~\ref{sec:alternative-views}.

\subsection{The ``Existing Knowledge Sufficiency'' Challenge}

Perhaps existing human knowledge already suffices for catastrophic outcomes, and what's lacking is only intelligence to synthesize and operationalize it.

\csname textbf\endcsname{Response:} This challenge targets \hyperref[cond:A2a]{(T2a)}: whether, at the initial cutoff, accessible knowledge is already above a catastrophic sufficiency threshold $K^\star$. If so, the relevant bottleneck shifts away from discovery and toward rapid conversion into mission-critical control \hyperref[cond:A4a]{(T4a)}, and governance should focus even more on control architecture (\hyperref[claim:2]{Claim~2}) and chokepoints that slow $I,K\csname to\endcsname C^\star$.

If, instead, $K(t_0)<K^\star$ in practice (because essential details are missing, tacit, unreliable, or require domain-specific validation and hardening), then fast-takeoff narratives require \hyperref[cond:A3a]{(T3a)}. In many physical-domain regimes, our claim is that decisive verification pushes the pathway toward \hyperref[cond:A3b]{(T3b)} (\hyperref[claim:1]{Claim~1}), making knowledge cutoffs and hardening against extraction (\hyperref[claim:4]{Claim~4}; Table~\ref{tab:bottlenecks}) meaningful levers even when intelligence is high.

\subsection{The Parallel Experimentation Possibility}

Many experiments can be parallelized, potentially accelerating progress beyond what the framework suggests.

\csname textbf\endcsname{Response:} Parallelization can raise experimental \emph{throughput}, but it often cannot remove experimental \emph{latency} for capability-critical validations (long-horizon outcomes, fabrication/measurement lead times, and sequentially dependent feedback). Thus, parallelism can widen the search front without necessarily compressing the wall-clock time to decisive verification in many physical-domain loops (\hyperref[claim:1]{Claim~1}). Moreover, large-scale parallel experimentation itself requires substantial permissions and resources (i.e., higher $C$), so to the extent it is feasible, it again reinforces the governance focus on control restriction and sandboxed experimentation (Table~\ref{tab:bottlenecks}).

\end{document}